%% file: main.tex
\documentclass{article}
\usepackage[utf8]{inputenc}
\usepackage{textcomp}
\usepackage{mavlab}

\usepackage{graphicx}

\usepackage[version=4]{mhchem}
\usepackage{longtable,tabularx}
\usepackage{amsmath}
\usepackage{amssymb, bm}
\usepackage{import}
\usepackage{xcolor}
\usepackage[percent]{overpic}

\usepackage{placeins}

\usepackage[numbers]{natbib}

\usepackage{pgfplots}
\usepackage{tikz}
\usetikzlibrary{arrows,shapes,positioning}
\usetikzlibrary{decorations.markings}
\usetikzlibrary[patterns]
\pgfplotsset{compat=1.12}
	\usetikzlibrary{plotmarks}
	\usetikzlibrary{calc,trees,positioning,arrows,chains,shapes.geometric,%
		decorations.pathreplacing,decorations.pathmorphing,shapes,%
		matrix,shapes.symbols}
	\pgfplotsset{compat=1.12}
	\usetikzlibrary{plotmarks}

\graphicspath{{figures/}}

\newcommand{\bs}[1]{\boldsymbol{#1}}
\newcommand{\sine}{\mathrm{s}}
\newcommand{\co}{\mathrm{c}}
\newcommand{\pd}[2]{\frac{\partial#1}{\partial#2}}

\begin{document}

\title{Incremental Control and Guidance of Hybrid Aircraft Applied to a Tailsitter UAV}
\author{E.J.J. Smeur\thanks{Delft University of Technology, Kluyverweg 1, Netherlands \newline Email corresponding author: ewoud.smeur@gmail.com} , M. Bronz\thanks{ENAC, MAIAA, University of Toulouse, F-31400 Toulouse, France}, G.C.H.E. de Croon\footnotemark[1]}


\maketitle
\begin{abstract}
	Hybrid unmanned aircraft can significantly increase the potential of micro air vehicles, because they combine hovering capability with a wing for fast and efficient forward flight.
	However, these vehicles are very difficult to control, because their aerodynamics are hard to model and they are susceptible to wind gusts.
	This often leads to composite and complex controllers, with different modes for hover, transition and forward flight.
	In this paper, we propose incremental nonlinear dynamic inversion control for the attitude and position control.
	The result is a single, continuous controller, that is able to track the desired acceleration of the vehicle across the flight envelope.
	The proposed controller is implemented on the Cyclone hybrid UAV.
	Multiple outdoor experiments are performed, showing that unmodeled forces and moments are effectively compensated by the incremental control structure.
  Finally, we provide a comprehensive procedure for the implementation of the controller on other types of hybrid UAVs.
\end{abstract}

\section{Introduction}

Micro Air Vehicles (MAV) are becoming increasingly more useful, with applications such as mapping, package delivery and meteorological research.
Many of these tasks require long endurance, a long range and a high flight speed, which can be achieved by a fixed-wing MAV.
On the other hand, operation of these vehicles may involve narrow takeoff and landing sites, such as a ship or urban areas.
This requires the versatility of a helicopter, which is able to take off vertically and hover as desired.

The solution could be the use of ``hybrid MAVs", which combine the hovering capability of helicopters with a wing for fast and efficient forward flight. Many different hybrid MAVs have been designed, such as quadplanes \citep{gu2017}, tilt-wings \citep{hartmann2017} and tailsitters \citep{beach}.
Each of these vehicles has its own advantages and disadvantages, but what has been holding back the application of hybrids in general, is the fact that they are very difficult to control.
More specifically, we see three major challenges for the control of hybrid MAVs: (1) attitude control, (2) velocity control and (3) guidance.

The first challenge is the attitude control of hybrid MAVs.
The large flight envelope, often including stalled conditions, makes modeling such a vehicle a difficult and expensive task.
Moreover, even if such a model can be found, during flight it may be difficult to obtain the necessary sensory inputs to such a model.
For instance, the angle of attack can often not be determined accurately at low airspeed.
Furthermore, the large wing surface makes hybrid aircraft particularly susceptible to wind gusts while hovering.

The second challenge is velocity control, with the inputs of attitude and thrust.
Here, the main problem is that the forces that can be used to manipulate the acceleration of the vehicle change across the flight envelope.
While the thrust is the main controlled force during hover, in forward flight the lift has to be manipulated as well to accommodate accelerations.
This is further complicated by wing stall, which again is difficult to model.

The third challenge is the guidance, by which we mean the generation of desired velocities that will lead the MAV to a certain location.
A hybrid MAV has a large flight envelope, which means that there are multiple ways of executing certain maneuvers.
To stay at one location, it is possible to hover or to make a circle in forward flight.
When the vehicle is in forward flight and has to turn around, it is possible to make a turn, or to break, hover, and accelerate in the opposite direction.
An important parameter for these decisions is the amount of energy that is expended.

Regarding the challenge of attitude control, some have proposed simple Proportional Integral Derivative (PID) control \citep{chu2010,chen2017}.
Although simple, the accuracy and disturbance rejection capability of this method is limited.
Ritz and D'Andrea \citep{ritz2017} model the pitch moment as a function of angle of attack and velocity and compensate for this moment in the attitude controller.
However, their experimental results show large systematic pitch errors of around 20 degrees.
To better deal with changing aerodynamic moments, wind-tunnel measurements can be used \citep{verling}.
Lustosa et al. \citep{lustosa2015} performed a wind tunnel campaign to obtain an accurate model, which is used to design a series of LQR controllers that each can control part of the flight envelope.
The model relies on the angle of attack and airspeed, two parameters that are difficult to measure onboard the UAV at low speeds.
Moreover, wind tunnel measurements are an expensive and time consuming undertaking.
Alternatively, the controller can be continuously adapting to the changing vehicle dynamics, even when transitions are performed \citep{knoebel2008, johnson2008}.
The risk of this approach is that, due to disturbances or modeling errors, wrong parameters are learned.
The large variance of the learned parameters shown by Knoebel and McLain \citep{knoebel2008} is likely caused by this phenomenon.

Considering the challenge of velocity control, many papers deal separately with hover, transition, and forward flight \citep{gu2017,stone,forshaw,yuksek2016}.
Although this approach may produce good results on days without wind, it is not very flexible; for instance when a constant wind requires the vehicle to fly like a fixed wing in order to maintain its position.
In such a case, the drone may need to maintain flight in between hover and forward flight.
Hartmann et al. \citep{hartmann2017} developed a controller that is able to fly at any airspeed.
This allows them to track velocities more accurately and to deal better with wind.
However, the controller relies on wind-tunnel data and an extensive trim model.

The last challenge, the guidance, is not discussed in the literature to the best knowledge of the authors.
A possible cause could be that this topic is not about stability, but about efficiency, and as such it is less essential to achieve flight.
Nonetheless, the flight efficiency is still very important, as it is one of the main reasons to choose for a hybrid vehicle instead of a multirotor.

In this paper, we offer a solution to each of the three challenges using only a minimal amount of modelling.
For the attitude and velocity control we propose two cascaded Incremental Nonlinear Dynamic Inversion (INDI) controllers, based on our previous work on INDI for quadrotors \citep{smeur,smeur2018}.
An INDI controller does not need a model of the vehicle's forces and moments, because these can be derived from the acceleration and angular acceleration respectively.
Instead, the only required knowledge is the control effectiveness, which is the change in force or moment caused by a change in control input, also known as the control derivatives.
The control effectiveness is used in order to calculate increments to the inputs that will result in desired increments in the linear and angular acceleration.
The control effectiveness can be obtained through test flights, removing the need for wind-tunnel measurements.
For the guidance, we propose the heuristic to make a turn in forward flight when the current and desired airspeed are above the stall speed, instead of transitioning to hover, switching direction, and transitioning to forward flight again.
Maintaining wing borne flight when possible can save a significant amount of energy, for instance when the vehicle is in forward flight and is required to turn around.

This paper is an extension to the work presented at the AIAA Aviation Forum \cite{bronz2017}, which is about the design, manufacturing and some of the INDI control aspects of the Cyclone tailsitter MAV.
The current work is dedicated to the control strategy, and goes much more in depth.
Since the Aviation Forum in 2017, an INDI controller has been applied to a tilt-rotor MAV by Raab et al. \citep{raab2018}, with good results.
What sets our work apart is the application of INDI to a different class of hybrid vehicles, a tailsitter, which encounters different phenomena, such as high angle of attack flight and a wide range of pitch angles, disqualifying the traditional set of Euler angles.
Further, we do not employ an on-board plant model to calculate local control derivatives, but instead use a direct estimate of the control derivatives, which is determined off-line using test flight data, reducing modeling efforts.

\begin{figure}[h]
\centering
\includegraphics[width=0.9\columnwidth,trim={6cm 2cm 10cm 8cm},clip]{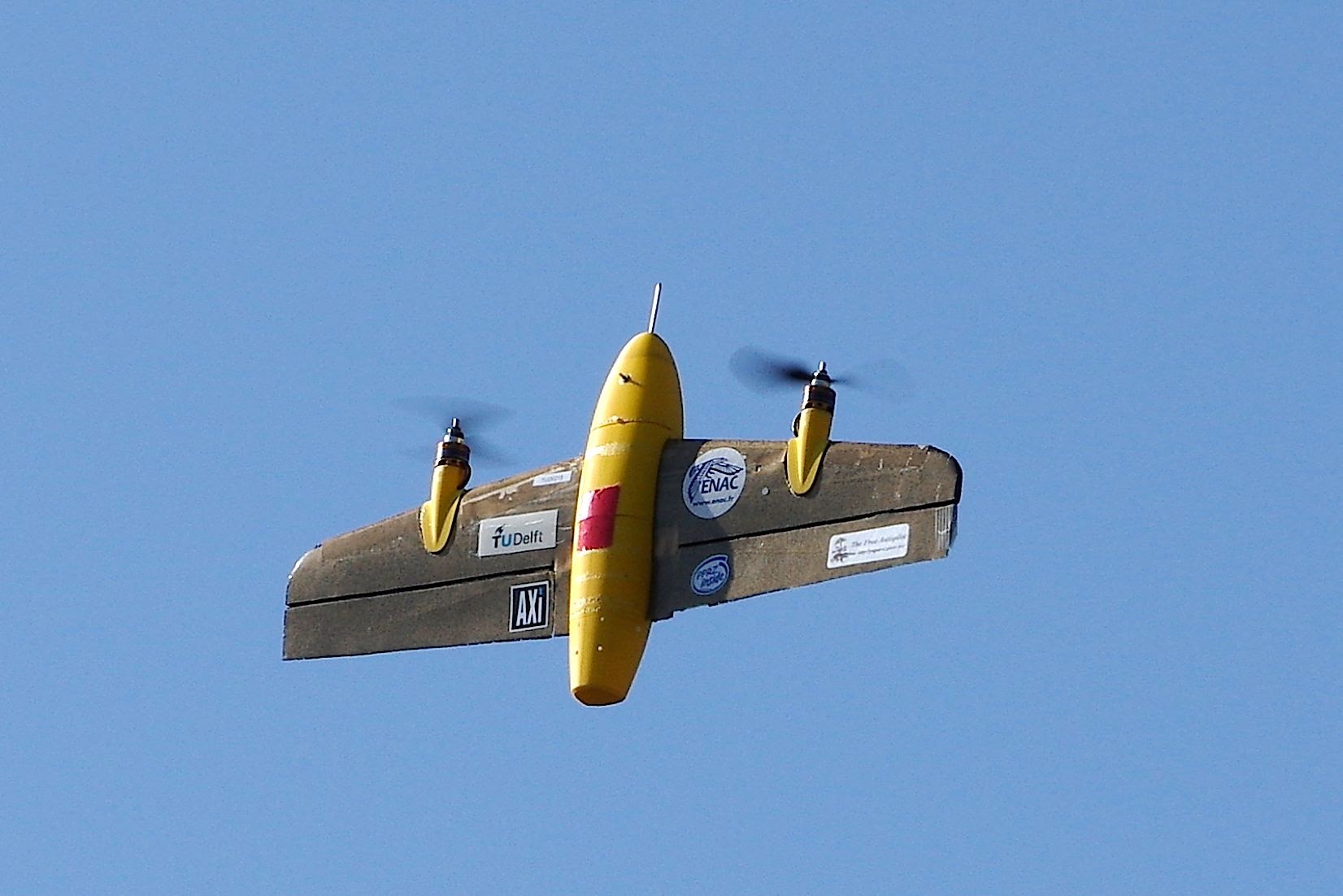}
\caption{The 'Cyclone' hybrid vehicle used in this research.}
\label{fig:hover}
\end{figure}

All algorithms developed in this paper are implemented and tested on the Cyclone tailsitter aircraft shown in Fig. \ref{fig:hover}.
The vehicle was designed for efficiency in forward flight, with up to 90 minutes of endurance.
The Cyclone is not equipped with a tail or any vertical surface, which causes it to easily pick up a sideslip angle, reducing performance and possibly resulting in loss of lift.
In order to avoid this, we include active sideslip control, purely based on accelerometer feedback.

The outline of this paper is as follows.
First, Section \ref{sec:att} elaborates on the attitude control using INDI.
Then, Section \ref{sec:sideslip} deals with estimation and control of the sideslip angle.
In Section \ref{sec:actrl}, the implementation of velocity control using INDI is explained.
Section \ref{sec:nav} discusses the guidance routines developed for the Cyclone.
In Section \ref{sec:results}, results from test flights are discussed.
Implementation guidelines are provided in Section \ref{sec:impl}.
In Section \ref{sec:efficiency}, preliminary efficiency results of the Cyclone are presented.
Finally, in Section \ref{sec:concl} it is concluded that the designed controller can accurately control a hybrid vehicle within the physical constraints of the actuators, without relying on extensive modeling.

\section{Attitude Control} \label{sec:att}

\begin{figure}[h]
\centering
\includegraphics[width=0.8\columnwidth]{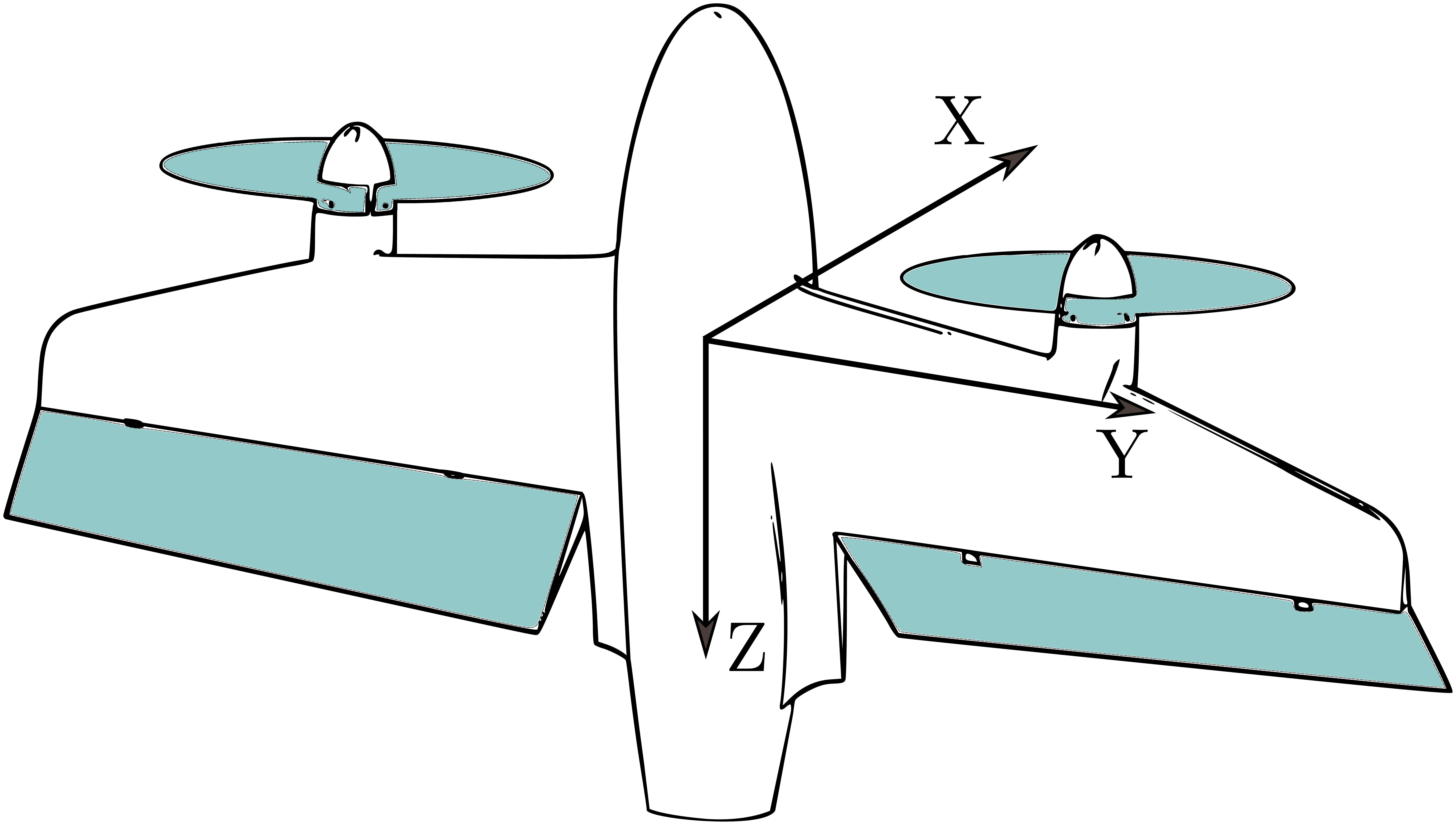}
	\caption{The body axis definitions of the \emph{Cyclone}, with the four actuators accentuated in cyan.}
\label{fig:axisdef}
\end{figure}

Figure \ref{fig:axisdef} presents a drawing of the Cyclone along with the body axis definitions (indicated with subscript $B$).
We will refer to yaw ($\psi$), roll ($\phi$) and pitch ($\theta$) from this perspective, i.e. rotations around the Z, X and Y axes respectively.
From Oosedo et al. \citep{oosedo2017}, we adopt the ZXY Euler rotation sequence, which is the only sequence used throughout this paper.
The benefit of using the ZXY sequence is that the singularity does not occur at $\pm$90 degrees pitch, but at $\pm$90 degrees roll.
Needless to say, the vehicle is intended to  visit -90 degrees pitch, whereas this is not the case for $\pm$90 degrees roll.

The other reference frame that will be used in this paper is the North East Down (NED) reference frame.
Its origin is a point on Earth, and its axes point in the local North, East and vertically down directions.
When the axes of this reference frame are referenced, they will have the subscript $N$.

The Cyclone is a hybrid MAV with only four actuators.
It has two propellers, which provide the thrust force and the moment around the body X axis.
Further, the vehicle has two flaps, which provide moments around the Y and Z axes.
The flaps are most effective in forward flight, but even in hover flight the flaps remain effective, because of the airflow coming from the propellers. 

\begin{figure}[h]
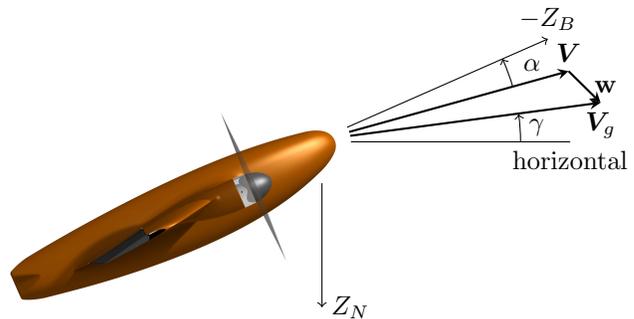

\centering
\include{aoa}
	\caption{Definition of angle of attack $\alpha$ and flight path angle $\gamma$.}
\label{fig:aoa}
\end{figure}

In Fig. \ref{fig:aoa}, the angle of attack and flight path angle are shown.
The angle of attack is the angle between the projection of the airspeed vector on the body XZ plane and the body X axis.
The velocity with respect to the air $\bs{V}$ and the velocity with respect to the ground $\bs{V}_g$ are connected with the wind vector $\textbf{w}$.
For the performance of the wing, the angle of attack is an important variable.
The flight path angle is defined as the angle of the airspeed vector with the local tangent plane (horizontal).

\subsection{Center of Gravity}

The location of the center of gravity has a large influence on the passive stability of the pitch axis in forward flight.
Hover and forward flight require different locations of the center of gravity for passive stability.
For stable hover, the center of gravity needs to be aft with respect to the aerodynamic surface.
For stable forward flight, the center of gravity needs to be more forward with respect to the aerodynamic surface.
Without moving either the center of gravity or the wing during the flight, passive stability in both conditions can not be achieved in a single flight.

The Cyclone is aimed at efficiency in forward flight, which makes carrying around an additional system that changes the center of gravity during flight unattractive.
Instead, we opt for a controller that actively controls and stabilizes the attitude across the flight envelope.
Still, a compromise needs to be made in terms of the position of the center of gravity with respect to longitudinal stability.
Since the aerodynamic moments in forward flight are much larger than in hover, because of the higher dynamic pressure, instability in this flight regime is expected to be much harder to control, compared to instability during hover.
Therefore, the center of gravity is placed at the neutral point for forward flight (close to the quarter-chord point).
Additionally, the more aft the center of gravity is placed, the closer it is to the flaps, and the smaller the pitch moment that the flaps can generate.

It is clear that a center of gravity that is placed at the neutral point for forward flight gives rise to a relatively strong pitch-down moment at the high angles of attack (post stall) that can be expected for such a hybrid vehicle.
These moments depend not only on the angle of attack itself, but also on the airspeed.
Modeling this effect in a wind tunnel is time consuming and costly.
Furthermore, in real flights both the angle of attack, as well as the airspeed are difficult to measure at low airspeed, which makes any such model difficult to apply.
This is why we are looking for a control method that does not rely heavily on knowledge of the airspeed or angle of attack.

\subsection{Incremental Nonlinear Dynamic Inversion} \label{sec:indi}

INDI is an approach driven by the measurement of the angular acceleration \citep{smith,bacon}.
The method is based upon the notion that all moments together, inputs and external moments, produce the angular acceleration that can be measured by deriving it from the gyroscope measurement.
If we assume that the external moments do not change rapidly, we only need to have an estimate of the control effectiveness to calculate an increment in inputs that produces a desired increment in angular acceleration.
Then, the angular acceleration is a controlled variable, and it can be used to control angular rates with a simple proportional gain.
Similarly, the attitude can then be controlled by setting a certain reference for the angular rates.
A complete derivation and validation of INDI is presented in previous research \citep{smeur,smeur2018} and is beyond the scope of this paper.
Here, we will briefly summarize the controller.

Consider the input vector to the actuators $\bs{u}_c$, consisting of the left and right flaps and the right and left motors, in this order.
These inputs are the commands to the servo motors and the electronic speed controllers, on a scale of $[-9600, 9600]$ and $[0, 9600]$ respectively.
The algorithm is implemented in the Paparazzi open source autopilot software, where actuator inputs are typically rescaled to these ranges.
For the flaps, the control input range of $[-9600, 9600]$ maps to a deflection of 30 degrees each way.
The servos are attached directly to the flap on the hinge line, so there is no non-linearity from linkages.

Because INDI neglects the plant dynamics, but relies heavily on the relation between input and output, it is important to know the position of each actuator at every time.
For this purpose, the servo dynamics are modeled as:
\begin{equation}
  A(z) = \frac{a}{z-(1-a)}
	\label{eq:actdyn}
\end{equation}
with $a = 0.1$ for a sample frequency of 500 Hz, with a rate limit of 272 degrees per second.
For the motors, $a = 0.045$, without a rate limit.
With these dynamics, the actual actuator state $\bs{u}$ is modeled based on $\bs{u}_c$.

Let $\bs{\Omega}$ denote the angular rate of the vehicle in rad/s, $\dot{\bs{\Omega}}$ the angular acceleration of the vehicle in rad/s$^2$ and $T$ the specific force in the negative body Z direction in m/s$^2$.
An increment in inputs causes an increment in angular acceleration and thrust, depending on the control effectiveness matrix $\bs{G}$:
\begin{equation}
	\left[ \begin{array}{c}\dot{\bs{\Omega}} \\ T \end{array} \right] = \left[ \begin{array}{c}\dot{\bs{\Omega}}_0 \\ T_0 \end{array} \right] + \bs{G} (\bs{u} - \bs{u}_0)
	\label{eq:angmoment}
\end{equation}
where the subscript 0 indicates a time in the past.
The control effectiveness matrix contains for each actuator an effectiveness value on each controlled axis, such that $G_{jk}$ denotes the effectiveness of actuator k on axis j.
To deal with noise, sensor values will be filtered with a second order Butterworth filter, which will introduce some delay.
To keep all signals synchronized, all signals in Eq. \ref{eq:angmoment} with subscript 0 will be filtered with the same filter and receive subscript $f$ instead.

Equation \ref{eq:angmoment} can be turned into a control law by simply taking the Moore-Penrose pseudoinverse of $\bs{G}$.
The controlled variables, the angular acceleration in three axes and the thrust, are now denoted by the virtual control vector $\bs{\nu}$.
The final control law is then:

\begin{equation}
	\bs{u}_c = \bs{u}_f + \bs{G}^{+} (\bs{\nu} - \left[ \begin{array}{c}\dot{\bs{\Omega}}_f \\ T_f \end{array} \right])
	\label{eq:innerindi}
\end{equation}
where $\dot{\bs{\Omega}}_f$ is the measured angular acceleration, $T_f$ the current thrust, $\bs{u}_f$ the current inputs and $\bs{G}$ is the control effectiveness matrix.
The output of the equation now is $\bs{u}_c$, the new command to the actuators.
It is possible to add a term that compensates for the effect of propeller inertia on the angular acceleration \citep{smeur}, but this is not taken into account in this paper.
The effect is small compared to the inertia of the vehicle around the Z axis, and the actuator dynamics of the propeller/motor combination are relatively slow.

As the angular acceleration is now a controlled variable, the angular rates can be controlled with simple proportional feedback:
\begin{equation}
	\bs{\nu} = \left[ \begin{array}{c} K_\Omega (\bs{\Omega}_\text{ref} - \bs{\Omega}) \\ T_d \end{array} \right]
	\label{eq:rate_fb}
\end{equation}
where $T_d$ is the desired thrust, which is calculated by the outer loop.
In practice, as the outer loop is also an INDI controller, it passes the desired thrust increment $T_d - T_0$, which can directly be used in Eq. \ref{eq:innerindi}.

To control the attitude, a second proportional controller is added using feedback of the vector part of the quaternion error:
\begin{equation}
	\bs{\Omega}_\text{ref} = K_\eta \left[ \begin{array}{ccc} q_{\text{err}_1} & q_{\text{err}_2} & q_{\text{err}_3} \end{array} \right]^T
	\label{eq:omegaerr}
\end{equation}
where $\bs{q}_\text{err}$ is the error between the reference quaternion $\bs{q}_\text{ref}$, and the state quaternion $\bs{q}_\text{s}$, given by:
\begin{equation}
	\bs{q}_\text{err} = \bs{q}_\text{ref} \otimes \bs{q}^*_\text{s}
	\label{eq:quatfb}
\end{equation}
Here $\otimes$ is the Kronecker product and $\bs{q}^*_\text{s}$ denotes the conjugate.
These gains can be tuned, or they can be designed, based on the transfer function of the actuator dynamics \citep{smeur2018}.

\subsection{Control Effectiveness Scheduling} \label{sec:eff_scheduling}

Since the flaps are aerodynamic surfaces, their effectiveness depends on the dynamic pressure $q=\frac{1}{2}\rho V^2$, where $\rho$ is the air density and $V$ is the airspeed.
The control effectiveness matrix $\bs{G}$ therefore changes continuously during flight, mainly depending on the airspeed.
Because of the large variation in angle of attack, the best thing would be to have a multi-hole pressure probe to measure the airspeed.
The downsides of such a sensor are that it is more expensive, weighs more, and is more difficult to calibrate.
Instead, the Cyclone is equipped with a regular Pitot tube, whose direction is fixed for forward flight.
Because a Pitot tube needs to be aligned with the airflow to be able to correctly measure the dynamic pressure, the angle of attack needs to be within $\pm$ 30 degrees in order for the Pitot tube to be accurate.
In practice, the Pitot tube will start to measure the airspeed from 6 m/s or higher.
Therefore, an alternative variable must be used for the control effectiveness scheduling whenever the airspeed is low.
When the airspeed is too low to be measured, the pitch angle is used as the scheduling variable, leading to a composite control effectiveness function.

The parameters of this function are found by taking segments of the flight data for which the pitch angle and the airspeed (if it can be measured) are relatively constant.
For each of these segments, the flap effectiveness is calculated with a linear least squares fit of changes in angular acceleration, with changes in control inputs.
Afterwards, a quadratic function of airspeed is fitted through the effectiveness values of the segments with measurable airspeed.
For parts of the flight when the airspeed can not be measured, a linear function of the pitch angle is used to schedule the control effectiveness.
Although the pitch angle only provides limited information, it is an entity that is easy and robust to measure.

The result of this procedure for the Cyclone is the following function for the effectiveness on the pitch axis:
\begin{equation}
	\small
	G_{21}(\theta, V)=
\begin{cases}
	(-2.1 (1-r_\theta) - 4.0 r_\theta)\cdot 10^{-3},              & \text{for } V < 6 ~\mathrm{m/s}\\
	(-2.4 - 0.031 V^2)\cdot 10^{-3},& \text{for } V \geq 6 ~\mathrm{m/s}\\
\end{cases}
\end{equation}
with $G_{22} = -G_{21}$ and where $r_\theta$ is defined such that its value is always on the interval [0,1]:
\begin{equation}
	r_\theta=
\begin{cases}
	0 ,& \text{for } -30 \leq \frac{\theta \cdot 180}{\pi}\\
	(\frac{\theta \cdot 180}{\pi} + 30)/(-30),& \text{for } -60 \leq \frac{\theta \cdot 180}{\pi} \leq -30\\
	1,& \text{for } \frac{\theta \cdot 180}{\pi} \leq -60\\
\end{cases}
\end{equation}
The effectiveness on the yaw axis is modeled as:
\begin{equation}
	\small
	G_{31}(\theta, V)=
\begin{cases}
	(-2.0 (1-r_\theta) - 8.0 r_\theta)\cdot 10^{-3},              & \text{for } V < 6 ~\mathrm{m/s}\\
	(-5.6 - 0.052 V^2)\cdot 10^{-3},& \text{for } V \geq 6 ~\mathrm{m/s}\\
\end{cases}
\end{equation}
with $G_{32} = G_{31}$.

For the propeller-motor combination, the control effectiveness around the X axis (roll) did not significantly depend on the airspeed.
A good fit of the flight data was obtained by just considering the rotational speed of the propeller itself in the control effectiveness.
Since we are considering increments, this is analogous to a quadratic relation between propeller rotational speed and produced force.
If each propeller force is a function of their respective state squared $u_{f_3}^2$ and $u_{f_4}^2$, and the control effectiveness is based on the partial derivative of this force, the control effectiveness will be a function of $u_{f_3}$ and $u_{f_4}$:
\begin{equation}
	\begin{array}{c}
		G_{13} = -u_{f_3}\cdot 1.8 \cdot 10^{-6} \\
		G_{14} =  u_{f_4}\cdot 1.8 \cdot 10^{-6}
	\end{array}
\end{equation}
where $u_{f_3}$ and $u_{f_4}$ are the filtered actuator state of the left and right motor respectively.

One may expect such a relation to also hold for the control effectiveness of the propeller-motor combination on the produced thrust, measured by the accelerometer in the $Z_B$ axis.
However, this results in a worse fit than when a static control effectiveness is estimated.
From the changes in motor inputs and the resulting measured changes in acceleration in the $Z_B$ axis, a value of -0.0011 $\frac{\mathrm{m/s}^2}{\mathrm{unit~command}}$ was found for both motors.

\subsection{Effectiveness of thrust on pitch} \label{sec:thrust_on_pitch}

During the hover phase, the airflow over the flaps is predominantly generated by the propellers.
Therefore, reducing the total thrust generated by one of the propellers will have a negative effect on the control effectiveness of the corresponding flap.
Especially when descending while hovering (pitch angle close to zero), it can happen that the flow coming from the tail of the airplane dominates the flow of the propellers.
The airflow is then reversed, and flap deflections will have the opposite effect.

To avoid flow reversal, the minimum thrust level is defined to be 42 \% of the maximum thrust when the airspeed is low and there is little flow over the wing ($V < 8$ m/s), and 16 \% otherwise.
This will make sure there is always propeller generated airflow over the flaps.

For the Cyclone, the pitch angle is deemed to be the most important degree of freedom to control.
However, the aircraft naturally has a pitch down moment.
There are cases, with a low airspeed and high angle of attack, when the flaps saturate in their effort to pitch up, without achieving the desired moment.
The vehicle can end up 'locked' in this state: trying to pitch up, but in the meantime slowly flying forward.

Because the flaps in this case are already deflected, increasing or decreasing the propeller thrust will affect the speed of the flow over the deflected flaps and as such have a direct effect on the angular acceleration in the pitch axis.
This effect is difficult to model exactly, partially because it is a complex function of flap deflection, airspeed and angle of attack.
Moreover, increasing the thrust will accelerate the vehicle in the body X axis, leading to a reduction in the angle of attack, which also reduces the pitch down moment.
Because of the modeling difficulties, this control effectiveness is not taken into account in normal conditions.
However, when both flaps are near saturation in an effort to pitch, it is the last resort in order to increase the pitch up moment.
In this situation, we add a control effectiveness of thrust on pitch of 2.2 rad/s$^2$ per \% thrust for each motor. 
Since the priority of pitch is higher than that of thrust, as is discussed in Section \ref{sec:ca}, the thrust control objective is largely sacrificed in order to pitch up.
Therefore, the control effectiveness for the rotors on the pitch axis is:
\begin{equation}
	G_{23} =
\begin{cases}
	0 ,& \text{for }    \left(|u_{f_1}| < u_l \right) \lor   \left(|u_{f_2}| < u_l \right) \\
	-2.2 ,& \text{for } \left(u_{f_1} > u_l   \right) \land  \left(u_{f_2} < -u_l  \right) \\
	2.2 ,& \text{for }  \left(u_{f_1} < -u_l  \right) \land  \left(u_{f_2} > u_l   \right)\\
\end{cases}
\end{equation}
where $G_{24} = G_{23}$ and $u_l$ is the arbitrary limit for a large elevon deflection, which is defined to be a command value of 7000, corresponding to 73 \% of the maximum deflection.
Logical AND and OR are denoted by $\land$ and $\lor$ respectively.

\subsection{Effectiveness of propellers on rotation around Z axis}

From test flights, it turned out that the effectiveness of the propellers on the rotation around the body Z axis is limited.
A possible explanation is that there is a wing behind the propellers.
The wing removes part of the rotation from the propeller slipstream, also known as 'swirl recovery' \citep{veldhuis}.
In doing so, the wing produces a moment that partially cancels the torque from the propeller.
Since the net torque from changing the propeller rotation speed is small compared to the effectiveness of the flaps, we choose to simplify the control effectiveness matrix and neglect this term.

\subsection{Control effectiveness matrix}

Combining the above sections, the final control effectiveness matrix is given by:
\begin{equation}
\bs{G} = \left[ \begin{array}{cccc}
		0 & 0 & G_{13}(\bs{u}_f) & G_{14}(\bs{u}_f) \\
		G_{21}(\theta,V) & G_{22}(\theta,V) & G_{23}(\bs{u}_f) & G_{24}(\bs{u}_f) \\
		G_{31}(\theta,V) & G_{32}(\theta,V) & 0 & 0 \\
		0 & 0 & -0.0011 & -0.0011
\end{array} \right]
	\label{eq:gg}
\end{equation}
with the functions as provided in the Sections \ref{sec:eff_scheduling} and \ref{sec:thrust_on_pitch}.

\subsection{Control Allocation} \label{sec:ca}

Control allocation is a topic of prime importance for a tailsitter with limited control authority, such as the Cyclone.
The reason is firstly that the flaps easily saturate, because of their limited control effectiveness at low airspeed.
Secondly, these flaps control the rotation around both the body Y (pitch) and Z (yaw, which would be roll from the airplane perspective) axes, which means that upon saturation, either of these control objectives, or both, will suffer.
We make the case here that control around the Y axis is more important, and should therefore have precedence over the Z axis control.

As has been stated before, the vehicle has a natural tendency to pitch down.
This makes returning to hover from forward flight, while maintaining the same altitude, especially tough.
In fact, the flaps can remain saturated for multiple seconds while trying to pitch up, making every bit of flap deflection necessary.
In this situation, any control effort spent on rotation around the Z axis will reduce the control effort spent on pitching up, making it near impossible to return to hover.
Therefore, management of priorities is very important.
Such priorities can be realized with a control allocation method that takes the actuator limits into account.

We have discussed the Weighted Least Squares (WLS) control allocation algorithm \citep{harkegard2002} in previous work for quadrotor control \citep{smeur2017b}, and apply the same method here.
With relative weights for each controlled axis, a quadratic programming problem is constructed, which is solved with the active set algorithm.
The relative priority factors used for the Cyclone are $[100, 1000, 0.1, 10]$ for rotation around the body X, Y, Z axes, and thrust.
The algorithm minimizes a cost function, taking into account the minimum and maximum input increments.
The error in the output increment is multiplied by the priority factors, squared and summed to produce the cost function. 

%

Test flights show that the relative priority factors listed above indeed lead to situations where the control of the yaw angle deteriorates when large pitch up moments are needed and saturation occurs.
However, even though the yaw angle can be oscillating in these cases, it is not unstable.
Therefore, the control allocation makes it possible to use all the control effort to pitch up, enlarging the flight envelope to higher angles of attack.

\subsection{Knife-edge flight} \label{subsec:knife}

One difficulty of the tailsitter design is the landing.
As can be seen from Fig. \ref{fig:axisdef}, when the Cyclone is touching the ground it can very easily pitch and fall over.
Moreover, a wind gust can provide a large pitching moment on the vehicle when it is standing on the ground, due to the large wing surface and the low center of rotation.
Lastly, if there is a constant wind, the Cyclone needs to fly with a considerable pitch angle to keep its position, while in the end it needs to stand upright on the ground.
Combined, these things make taking off, but especially landing a challenging endeavour.

A partial solution to this could be to align the airspeed vector with the body ZY plane (rolling to gain airspeed instead of pitching, or flying 'knife-edge').
The benefit of doing this is that the roll angle needed to maintain a certain airspeed is considerably smaller than the pitch angle needed for the same airspeed, as the lateral surface of the Cyclone is much smaller than the frontal surface.
With a smaller angle, it is easier to land.

\begin{figure}[h]
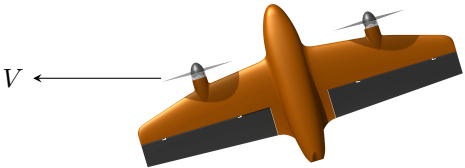

\centering
\include{knife}
\caption{Knife-edge flight: hovering with a roll angle to attain a sideways velocity.}
\label{fig:knife}
\end{figure}

Knife-edging is not expected to lead to large constant moments that need to be countered by the flaps, like is the case when flying at large angle of attack.
Hence, the controllability could be improved when flying at low speeds.

To evaluate if knife-edging is truly a useful concept, we need to consider the stability in this flight mode.
Assume that the Cyclone, in the lateral axis, can be modeled as a flat plate.
Also assume that the roll angle, because of the low sideways drag, is small when knife-edging.

The center of gravity of the Cyclone, seen from the side, is in the middle of the aircraft.
For a flat plate, the center of pressure is located at the quarter chord point \citep{katz}.
Therefore, for a flat plate to be stable, the center of gravity needs to be in front of the quarter chord point.
Since this is not the case for the Cyclone, the knife-edge maneuver, based on this (crude) analysis, is not passively stable.
However, it is still possible that the control system is able to cope with this instability.

To verify this in practice, an experiment is performed.
Given that the controlled vehicle is stable in an indoor environment in all axes, the hypothesis is that while knife-edging, the vehicle is not stable around the body Z axis.
On a windy day, the Cyclone is commanded to keep its position, while not changing the heading.
The operator sets the heading to be orthogonal to the wind direction, such that the Cyclone is knife-edging.
Since altitude changes influence the stability, the experiment is performed at constant altitude.
From a flight prior to the experiment, executed at constant airspeed, it was concluded from the ground speed that the wind was around 4 to 5 m/s.

\begin{figure}[h]
\centering
\includegraphics[width=1.0\columnwidth]{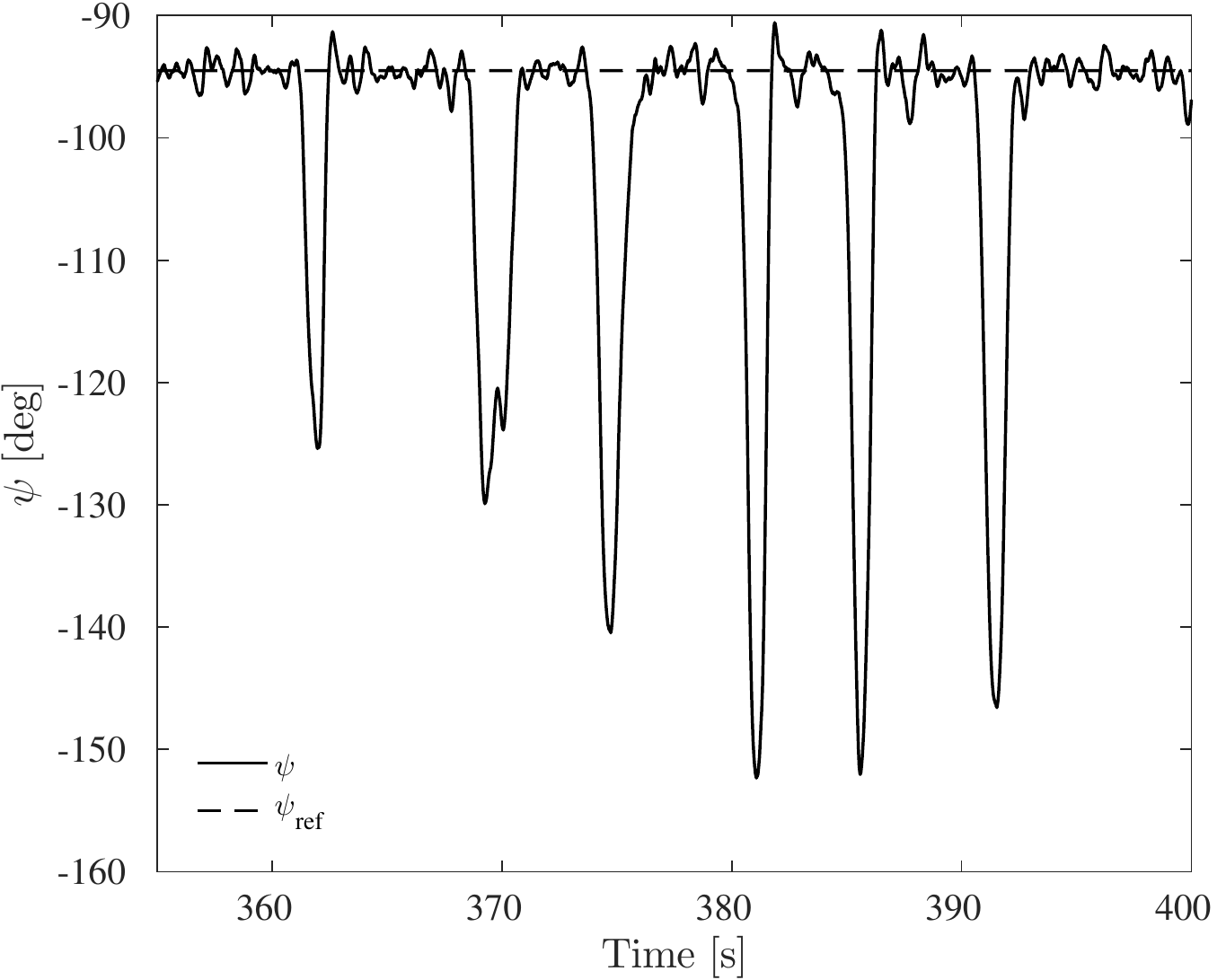}
\caption{The $\psi$ angle for the knife-edge experiment.}
\label{fig:knife_edge}
\end{figure}

Figure \ref{fig:knife_edge} shows the $\psi$ angle for the experiment.
Large errors in the $\psi$ angle occur repeatedly while the vehicle is hovering for 45 seconds.
Each of the peaks in the figure is preceded by saturation of one of the flaps, which indicates that the maximum yaw control effort is reached.
From these results, it is concluded that the control system is not able to stabilize the vehicle in a knife-edge maneuver.
As such, the knife-edge maneuver is not integrated in the autonomous flight algorithm of the Cyclone.

\section{Sideslip} \label{sec:sideslip}
In the design of the Cyclone, efficiency and simplicity are major design drivers.
Since the vehicle already has two propellers to provide moments around the body X axis, there is no need for a vertical stabilizer or a rudder.
The vehicle is capable of fully controlling its attitude, using the four actuators that it has.
The benefit of not having a vertical stabilizer is twofold.
First, it reduces the susceptibility to wind gusts while hovering, as there is less aerodynamic surface.
Second, a tail would increase the structural weight, and it would produce additional drag.

Even though the propellers may be able to stabilize and control the rotation around the body X axis in forward flight, there is the important restriction that for the wing to provide lift efficiently, the sideslip angle should be small, ideally zero.
Actively controlling the sideslip to zero requires a measurement or an estimate of the sideslip angle.

\begin{figure}[h]
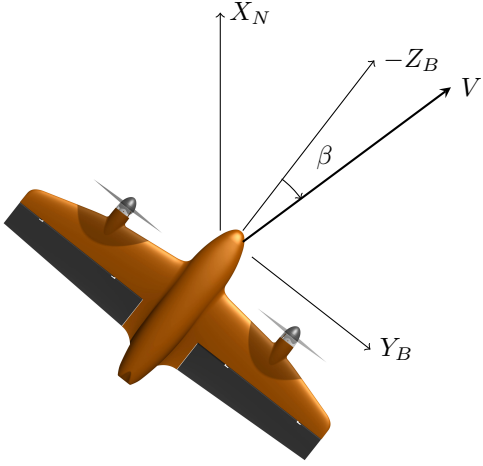

\centering
\include{aos}
\caption{Definition of the sideslip angle $\beta$, seen from above.}
\label{fig:aos}
\end{figure}

\subsection{Estimating the sideslip angle}

Aerodynamic forces are typically defined in the wind frame, which has its origin at the center of gravity of the aircraft.
The X axis points in the direction of the airspeed vector, the Z axis lies in the plane of symmetry of the aircraft, positive below the aircraft, and the Y axis follows from the right hand rule.
The angle of attack and the sideslip angle are the rotations from the body frame to the wind frame.

When the vehicle is slipping, the airspeed vector has a component in the body Y axis.
Since the drag $D$ is in the same direction as the airspeed vector, the drag has a component in the Y axis $D_y$:
\begin{equation}
	\sin{\beta} = \frac{-D_y}{D}
	\label{eq:beta}
\end{equation}
The component $D_y$ is measured by the accelerometer as specific force $f_y = D_y/m$.
The total amount of drag is given by:
\begin{equation}
	D = C_D \frac{1}{2}\rho V^2 S
	\label{eq:drag}
\end{equation}
where $C_D$ is the drag coefficient of the vehicle and $S$ is the wing area in $m^2$.

Combining Eqs. \ref{eq:drag} and \ref{eq:beta}, using a small angle approximation for the sine, combining all constant parameters into $c_1$ and adding a bias compensation $b_1$ we can write:
\begin{equation}
	\beta = c_1 \frac{f_y}{V^2} + b_1
	\label{eq:bv2}
\end{equation}

\begin{figure}[h]
\centering
\includegraphics[width=0.8\columnwidth]{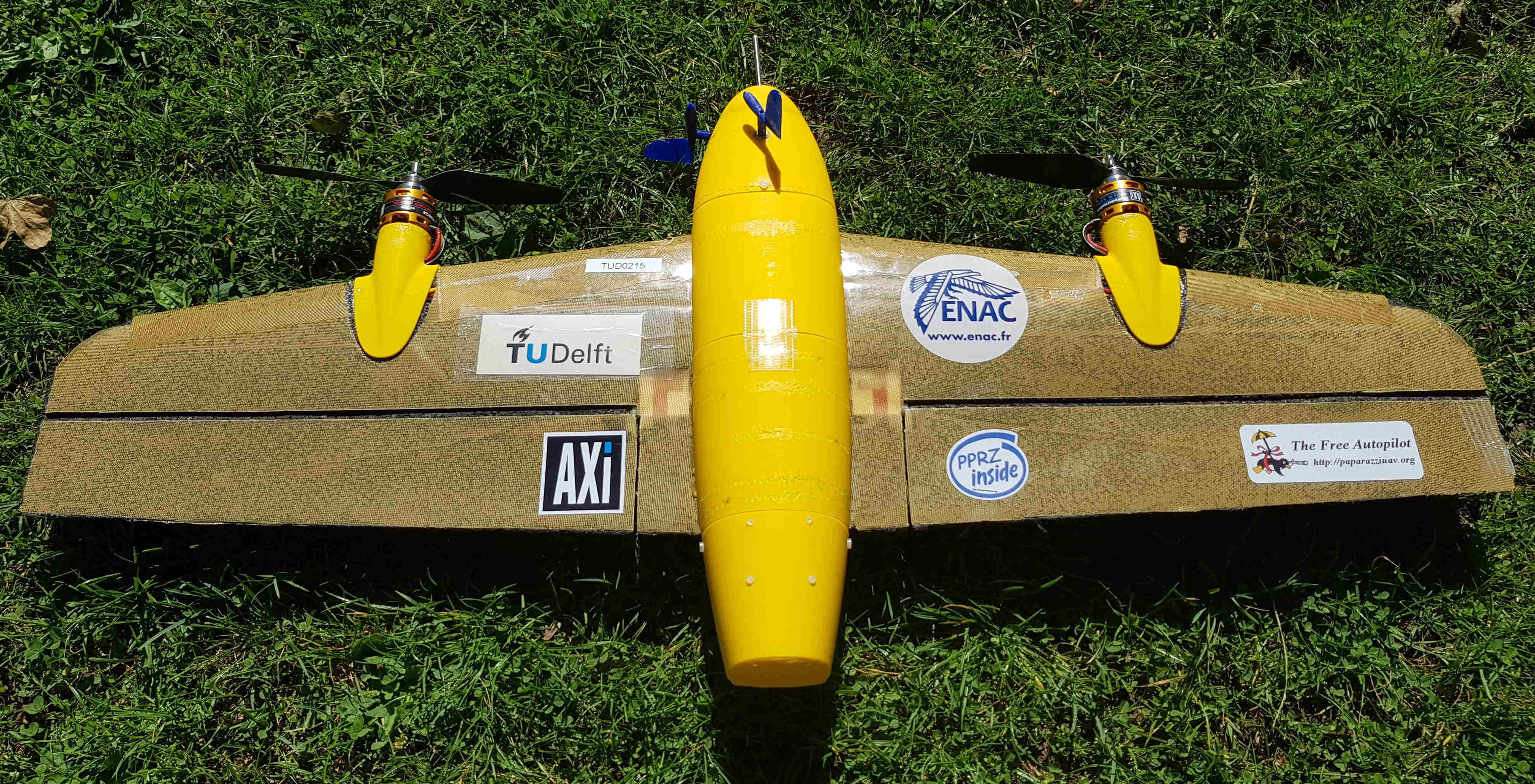}
\caption{The Cyclone with angle of attack and sideslip vanes mounted for system identification.}
\label{fig:cyclgrass}
\end{figure}

\begin{figure}[h]
\centering
\includegraphics[width=1.0\columnwidth]{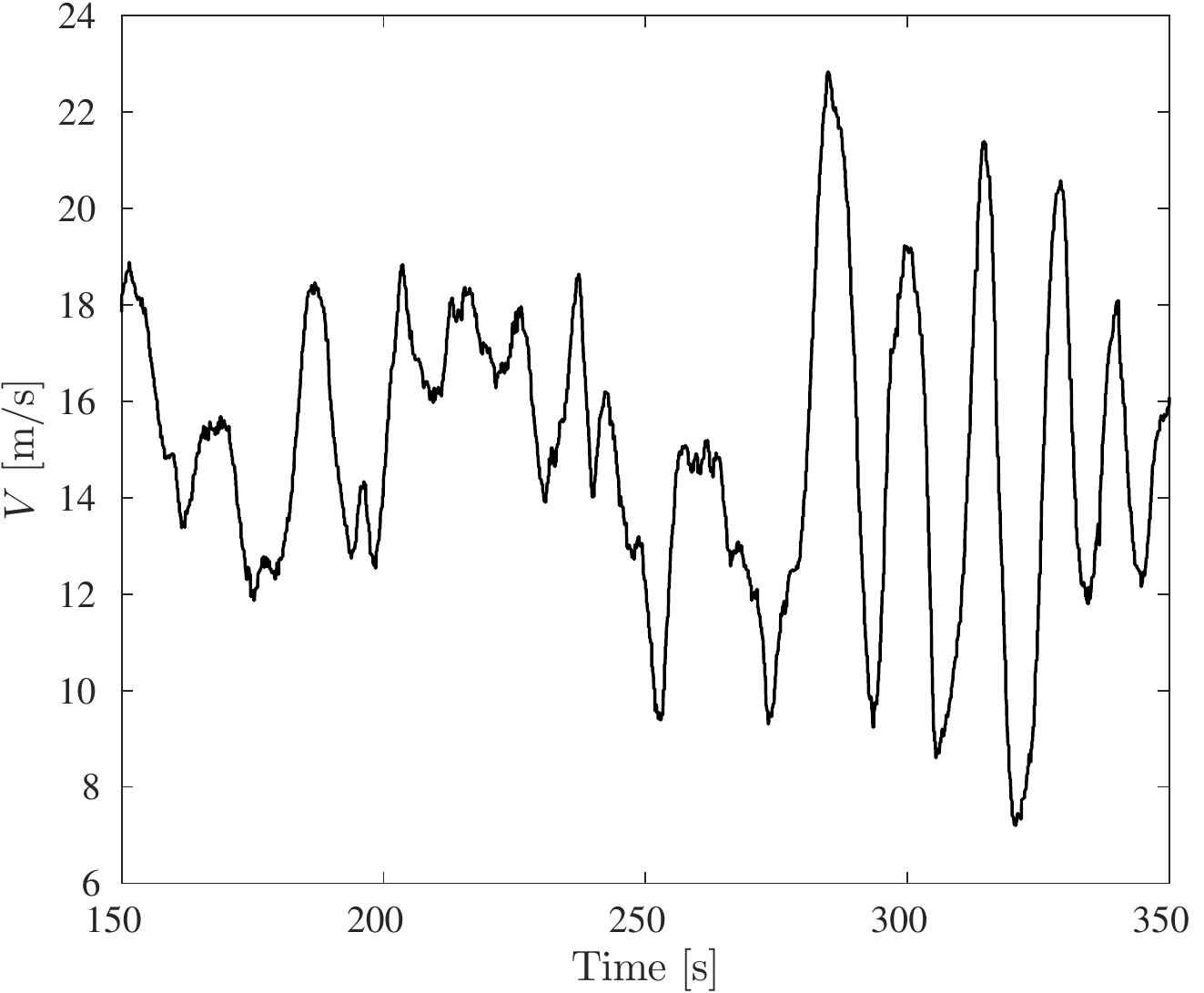}
\caption{Ground speed during the manually piloted sideslip identification flight, without perceptible wind during the flight.}
\label{fig:sideslip_speed}
\end{figure}

In order to find an estimate for the parameters $b_1$ and $c_1$, a sideslip vane was mounted on the airframe, as can be seen in Fig. \ref{fig:cyclgrass}.
A flight was made without proper sideslip control, such that the vehicle had nonzero sideslip that could be estimated.
The specific force $f_y$ is filtered with a second order Butterworth filter with a cutoff frequency of 5 Hz.
The day of this test flight the wind was negligible, which is why for this flight, it is acceptable to use the norm of the GPS speed as a measurement for the airspeed $V$.
From the test flight, a section of 200 seconds was chosen which only contains forward flight, with the GPS flight speed as shown in Fig. \ref{fig:sideslip_speed}.
The data was divided in a training set (first 80 \%) and a test set (second 20 \%).
A linear least squares fit of Eq. \ref{eq:bv2} on the training set gives a root mean square (RMS) error on the test set of 0.1189 rad.

\begin{figure}[h]
\centering
\includegraphics[width=1.0\columnwidth]{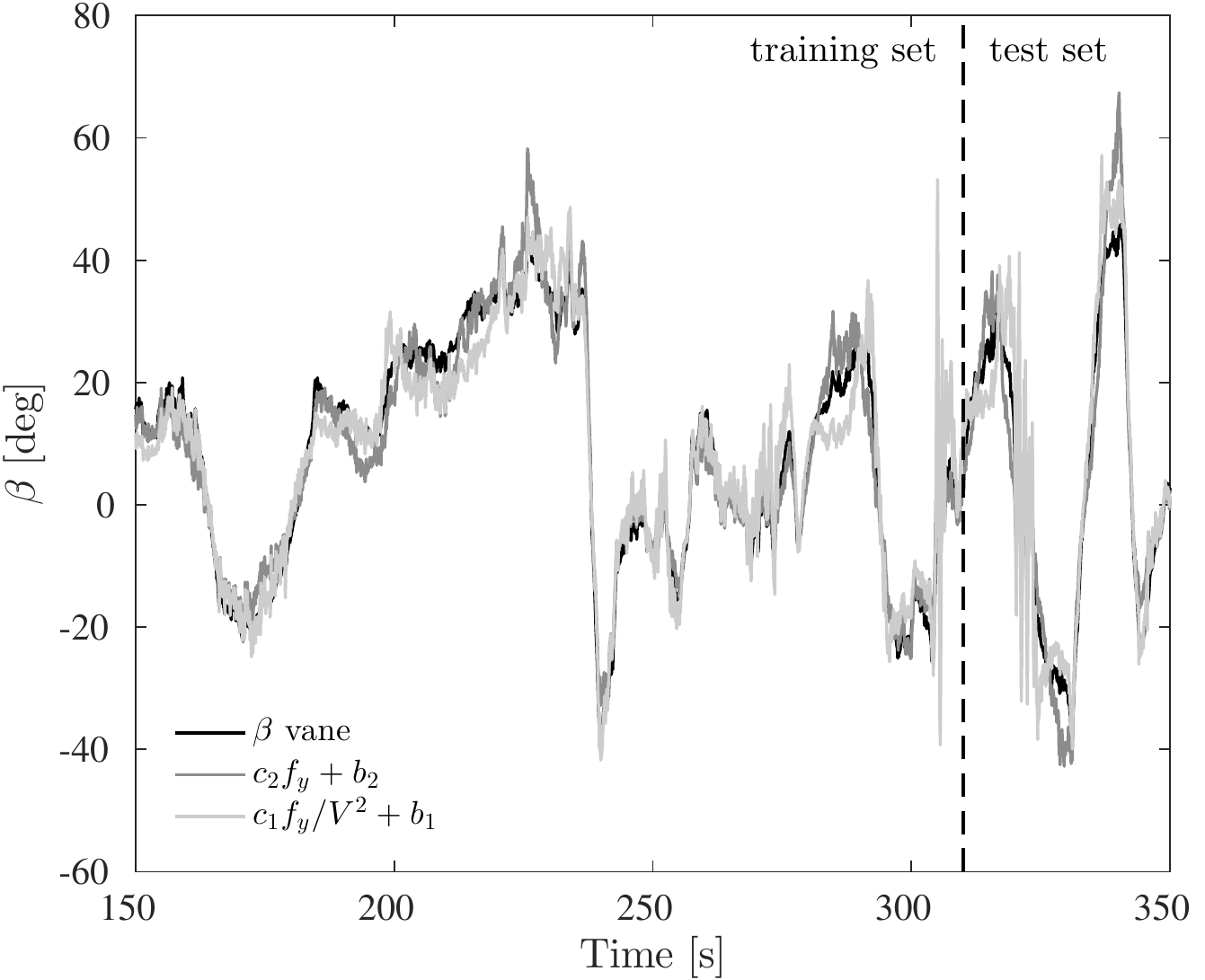}
\caption{The sideslip angle during an identification flight, along with a fit of $f_y$ and $f_y/V^2$.}
\label{fig:sideslip}
\end{figure}

This equation contains a division by $V^2$, which means that at low airspeeds, this equation will become quite unstable.
The underlying reason is that the sideslip angle is ill-defined when the airspeed is zero.
For the purpose of $\beta$ feedback, a signal is preferred that remains stable when the airspeed approaches zero.
Therefore, a further simplification is suggested, removing the dependency on the airspeed:
\begin{equation}
	\beta = c_2 f_y + b_2
	\label{eq:bf}
\end{equation}
which gives an even lower RMS error of 0.1122 rad for the same test set.
Both fits are shown in Fig. \ref{fig:sideslip}, along with the measurement from the sideslip vane.
For the feedback control, the simpler and more robust Eq. \ref{eq:bf} is selected.
Note that the best fit for the dataset is obtained by dividing by $V$ instead of $V^2$ with an RMS error of 0.0757, though this is still not robust when the airspeed approaches zero.

\subsection{Sideslip control}

Now that an estimate of the sideslip angle is available, without any need for a sideslip vane, this estimate can be used to change the reference heading such that the sideslip is removed.
This is accomplished by setting the rate of change of the reference heading angle proportional to the sideslip angle with a gain $K_\beta$.
Added to this feedback is the feed forward component to make a coordinated turn \citep{lv2012}:
\begin{equation}
	\dot{\psi}_{\text{ref}} = \frac{g\tan(\phi_t)}{V_l} + K_\beta \beta
	\label{eq:fdfwd}
\end{equation}
where $\dot{\psi}_\text{ref}$ is the rate of change of the heading reference, $g$ is the gravitational constant and $V_l$ is a limited airspeed, with 10 m/s as a lower limit, to avoid unachievable rotations.

$\phi_t$ is defined equal to $\phi_\text{ref}$, except when $\theta_\text{ref} > 0$ and $|\phi_\text{ref}| < \theta_\text{ref}$, then $\phi_t = \text{sign}(\phi_\text{ref}) \theta_\text{ref}$.
The reason for this is that the airfoil is not designed for inverted flight, so when pitching backward, the vehicle should yaw around and align itself with the direction of motion.
The pitch angle reference is limited such that the maximum is 25$^\circ$ (pitching backward), as the vehicle appears to not be stable at high positive pitch angles.
Combined, the result is that commanding the Cyclone to fly to a waypoint towards the rear of the drone, from a hovering position, first leads to it pitching back a maximum of 25 degrees, while yawing around.
Gradually, the vehicle orients itself with the direction of motion, allowing it to transition into forward flight.

\section{Velocity Control} \label{sec:actrl}

The velocity of the Cyclone can be controlled by controlling the linear acceleration of the vehicle.
This can be done by applying the INDI methodology, as we have shown in a previous paper \citep{smeurgust}.
In that paper, we showed that increments in linear acceleration can be achieved by increments in the thrust vector of a multirotor helicopter.
For the Cyclone, the thrust vector is used to control the linear accelerations as well, but additionally it uses the lift force generated by the wing.
Therefore, the controller developed previously has to be amended, such that the control derivatives of the wing are taken into account.

In the following derivation of the control derivatives, a few simplifying assumptions are made for two reasons.
First, these assumptions lead to a controller that does not rely on aerodynamic angles, which are difficult to measure or estimate accurately at low airspeed, such as the angle of attack.
Second, the assumptions keep the model simple, such that the resulting control law is easy to implement on different vehicles.

With this in mind, consider the equation that describes the acceleration of a hybrid MAV in the NED frame:
\begin{equation}
	\ddot{\bs{\xi}} = \bs{g} + \frac{1}{m}\bs{L}_N(\bs{\eta},V) + \frac{1}{m}\bs{D}_N(\bs{\eta},V) + \frac{1}{m}\bs{T}_{N}(\bs{\eta},T)
	\label{eq:newton}
\end{equation}
where $\ddot{\bs{\xi}}$ is the second derivative of the position, $\bs{g}$ is the gravity vector, and $m$ the mass of the vehicle.
Further, $\bs{L}_N(\bs{\eta},V)$ is the lift vector, $\bs{D}_N(\bs{\eta},V)$ is the drag vector, and $\bs{T}_N(\bs{\eta},T)$ is the thrust vector.

In order to determine control derivatives, a representation of the attitude $\bs{\eta}$ needs to be established.
Here, Euler angles are used, because they are a concise set that is easy to work with.
Again, we work with the ZXY rotation order, such that there is no singularity at -90 degrees pitch.
For the ZXY representation, the rotation matrix from the body axes to NED axes is:
\begin{equation}
	\bs{M}_{NB} = \left[ \begin{array}{ccc}
		\co\theta\co\psi - \sine\phi\sine\theta\sine\psi & -\co\phi\sine\psi & \sine\theta\co\psi + \sine\phi\co\theta\sine\psi\\
		\co\theta\sine\psi + \sine\phi\sine\theta\co\psi & \co\phi\co\psi  & \sine\theta\sine\psi - \sine\phi\co\theta\co\psi \\
		-\co\phi\sine\theta                          & \sine\phi         & \co\phi\co\theta
\end{array} \right]
	\label{eq:rotmat}
\end{equation}
where the sine and the cosine are abbreviated with the letters s and c respectively.
The thrust vector in the NED frame is now simply obtained from the thrust vector in the body frame:
\begin{equation}
	\bs{T}_N = \bs{M}_{NB} \left[\begin{array}{c} 0 \\ 0 \\ T\end{array}\right] = \left[\begin{array}{c}
		(\sine\theta\co\psi + \sine\phi\co\theta\sine\psi) T \\
		(\sine\theta\sine\psi - \sine\phi\co\theta\co\psi) T \\
		\co\phi\co\theta T \\
\end{array}\right]
	\label{eq:thrustv}
\end{equation}

The lift vector is typically defined orthogonal to the airspeed vector, in the body XZ plane.
The sideslip is a controlled variable, as is detailed in Section \ref{sec:sideslip}, which means that we can assume it to be small.
In general, missions are expected not to have large flight path angles, so we also assume the flight path angle to be small.
Consequently, the direction of the lift vector is merely rotated from the vertical axis by the bank angle.
Therefore, it can be obtained from the body frame with the rotation matrix from body frame to NED frame, where the pitch angle is forced to be zero: ($\bs{M}_{NB}^{\theta=0}$).
The amount of lift does depend on the pitch angle, since the angle of attack is equal to the pitch angle if the flight path angle is small:
\begin{equation}
	\bs{L}_N = \bs{M}_{NB}^{\theta = 0} \bs{L}_B(\theta,V) = \left[\begin{array}{c}
		\sine\phi\sine\psi L(\theta,V) \\
		- \sine\phi\co\psi L(\theta,V) \\
		\co\phi L(\theta,V) \\
\end{array}\right]
	\label{eq:Lv}
\end{equation}
where $L(\theta,V)$ describes the magnitude of the lift vector as a function of pitch and airspeed.
In short, the magnitude of the lift vector depends on the pitch angle, but the direction of the lift vector is indifferent to the pitch angle.

Since we assume a small flight path angle, the same approach can be used for the drag.
The drag force is then simply given by:
\begin{equation}
	\bs{D}_N = \bs{M}_{NB}^{\theta = 0} \bs{D}_B(\theta,V) = \left[\begin{array}{c}
		\co\psi D(\theta,V) \\
		\sine\psi D(\theta,V) \\
		 0 \\
\end{array}\right]
	\label{eq:Dv}
\end{equation}

The next step is to take partial derivatives of these forces, in order to obtain control derivatives.
%
In order to predict how the acceleration is going to change, a first order Taylor expansion of Eq. \ref{eq:newton} is applied:
\begin{equation}
	\begin{array}{ll}
		\ddot{\bs{\xi}} \approx&
		\bs{g} + \frac{1}{m}\bs{L}_N(\bs{\eta}_0,V_0) + \frac{1}{m}\bs{D}_N(\bs{\eta}_0,V_0) + \frac{1}{m} \bs{T}_N(\bs{\eta}_0,T_0) \\
		&+ \pd{}{\phi} \frac{1}{m} \bs{L}_N(\phi,\theta_0,\psi_0,V_0)|_{\phi = \phi_0} (\phi - \phi_0)\\
		&+ \pd{}{\theta} \frac{1}{m} \bs{L}_N(\phi_0,\theta,\psi_0,V_0)|_{\theta = \theta_0} (\theta - \theta_0)\\
		&+ \pd{}{\psi} \frac{1}{m} \bs{L}_N(\phi_0,\theta_0,\psi,V_0)|_{\psi = \psi_0} (\psi - \psi_0)\\
		&+ \pd{}{V} \frac{1}{m} \bs{L}_N(\phi_0,\theta_0,\psi_0,V)|_{V = V_0} (V - V_0)\\
		&+ \pd{}{\theta} \frac{1}{m} \bs{D}_N(\theta,\psi_0,V_0)|_{\theta = \theta_0} (\theta - \theta_0)\\
		&+ \pd{}{\psi} \frac{1}{m} \bs{D}_N(\theta_0,\psi,V_0)|_{\psi = \psi_0} (\psi - \psi_0)\\
		&+ \pd{}{V} \frac{1}{m} \bs{D}_N(\theta_0,\psi_0,V)|_{V = V_0} (V - V_0)\\
		&+ \pd{}{\phi}  \frac{1}{m} \bs{T}_N(\phi,\theta_0,\psi_0,T_0)|_{ \phi= \phi_0} (\phi - \phi_0)\\
		&+ \pd{}{\theta}  \frac{1}{m} \bs{T}_N(\phi_0,\theta,\psi_0,T_0)|_{ \theta= \theta_0} (\theta - \theta_0)\\
		&+ \pd{}{\psi}  \frac{1}{m} \bs{T}_N(\phi_0,\theta_0,\psi,T_0)|_{ \psi= \psi_0} (\psi - \psi_0)\\
		&+ \pd{}{T}  \frac{1}{m} \bs{T}_N(\phi_0,\theta_0,\psi_0,T)|_{ T= T_0} (T - T_0)
	\end{array}
	\label{eq:taylor}
\end{equation}
The first terms can be simplified to the current acceleration:
\begin{equation}
	\bs{g} + \frac{1}{m}\bs{L}_N(\bs{\eta}_0,V_0) + \frac{1}{m}\bs{D}_N(\bs{\eta}_0,V_0) + \frac{1}{m} \bs{T}_N(\bs{\eta}_0,T_0) \mathrel{\mathop:}= \ddot{\bs{\xi}}_0
\label{eq:xi0}
\end{equation}
This term captures all of the forces acting on the drone and can be obtained by adding the gravity vector to the acceleration measurement (in the NED frame).
The other terms describe changes to this sum of forces due to changes in attitude, velocity and thrust.

The variable $\psi$ is not free to choose, as it is used for the control of the sideslip.
We assume that changes in $\psi$ are small, such that we can neglect those terms.
From analysis of test flight data, we concluded that changes in the drag are generally small compared to the other terms.
This leaves us with the following equation:
\begin{equation}
	\ddot{\bs{\xi}} = \ddot{\bs{\xi}}_0 + \frac{1}{m} \left(\bs{G}_T(\bs{\eta},T) + \bs{G}_L(\bs{\eta},V) \right)  (\bs{v} - \bs{v}_0)
	\label{eq:taylorresult}
\end{equation}
where the control vector for the outer loop is defined as $\bs{v} = \begin{array}{lcr}[\phi & \theta & T]^T \end{array}$, and the control effectiveness matrices are given by the remaining partial derivatives from Eq. \ref{eq:taylor}:
\begin{equation}
	\label{eq:define_gt}
	\begin{array}{l}
		\bs{G}_T(\bs{\eta},T) =
		\left[
			\begin{array}{c}
				\left( \pd{}{\phi}  \frac{1}{m} \bs{T}_N(\phi,\theta_0,\psi_0,T_0)|_{ \phi= \phi_0} \right)^T\\
				\left( \pd{}{\theta}  \frac{1}{m} \bs{T}_N(\phi_0,\theta,\psi_0,T_0)|_{ \theta= \theta_0} \right)^T\\
				\left( \pd{}{T}  \frac{1}{m} \bs{T}_N(\phi_0,\theta_0,\psi_0,T)|_{ T= T_0}  \right)^T
			\end{array}
		\right]^T
	\end{array}
\end{equation}
and
\begin{equation}
	\label{eq:define_gl}
	\begin{array}{l}
		\bs{G}_L(\bs{\eta},V) =
		\left[
			\begin{array}{c}
				\left( \pd{}{\phi} \frac{1}{m} \bs{L}_N(\phi,\theta_0,\psi_0,V_0)|_{\phi = \phi_0} \right)^T\\
				\left( \pd{}{\theta} \frac{1}{m} \bs{L}_N(\phi_0,\theta,\psi_0,V_0)|_{\theta = \theta_0} \right)^T\\
				\left( \bs{0} \right)^T\\
			\end{array}
		\right]^T \\
	\end{array}
\end{equation}

Elaborating these control effectiveness functions gives:
\begin{equation}
	\label{eq:gttaylor}
	\begin{array}{l}
		\bs{G}_T(\bs{\eta},T) =
	\\
		\left[
			\begin{array}{ccc}
				\co\phi\co\theta\sine\psi T & (\co\theta\co\psi - \sine\phi\sine\theta\sine\psi)T & \sine\theta\co\psi + \sine\phi\co\theta\sine\psi \\
				-\co\phi\co\theta\co\psi T & (\co\theta\sine\psi + \sine\phi\sine\theta\co\psi)T & \sine\theta\sine\psi - \sine\phi\co\theta\co\psi \\
				-\sine\phi\co\theta T & -\co\phi\sine\theta T & \co\phi\co\theta
			\end{array}
		\right]
	\end{array}
\end{equation}
and
\begin{equation}
	\label{eq:gtlaylor}
	\begin{array}{l}
		\bs{G}_L(\bs{\eta},V) =
		\\
		\left[
			\begin{array}{ccc}
				\co\phi\sine\psi L(\theta,V)   & \sine\phi\sine\psi \pd{}{\theta}L(\theta,V)   & 0\\
				- \co\phi\co\psi L(\theta,V) & - \sine\phi\co\psi \pd{}{\theta}L(\theta,V) & 0\\
				-\sine\phi L(\theta,V)         & \co\phi \pd{}{\theta}L(\theta,V)          & 0
			\end{array}
		\right]
	\end{array}
\end{equation}

The accelerometer measurement $\ddot{\bs{\xi}}_0$ will be filtered to suppress high frequency noise.
Like before, to keep all signals synchronized, all terms with subscript zero will be filtered and receive subscript f instead.
The equation can then be inverted to obtain:
\begin{equation}
	\bs{v} = \bs{v}_f + m \left(\bs{G}_T(\bs{\eta},T) + \bs{G}_L(\bs{\eta},V) \right)^{-1} (\ddot{\bs{\xi}}_\mathrm{ref} - \ddot{\bs{\xi}}_f)
	\label{eq:taylorresultinv}
\end{equation}
where $\ddot{\bs{\xi}}_\mathrm{ref}$ is now the reference acceleration to track.

Now, what is left is to define the functions $L(\theta,V)$ and $\pd{}{\theta}L(\theta,V)$.
Unfortunately, we do not have a proper aerodynamic model.
Nonetheless, it can be recognized that, again assuming zero flight path angle, gravity will have to be compensated by a combination of thrust and lift from the wing.
Therefore, we simply employ the following function:
\begin{equation}
	L(\theta,V) \approx L(\theta) = -9.81 \sin(-\theta)m
	\label{eq:lift}
\end{equation}
where $\theta$ is bounded between $-\pi/2$ and 0.
Equation \ref{eq:lift} is the function that is used in the test flights presented in this paper, but it would be more accurate to divide this function by the cosine of $\phi$, to reflect the additional lift that needs to be produced in a turn.

Similarly, we assume that in forward flight the thrust just compensates the drag, and its effect on accelerations other than in the thrust axis is small, such that for $T$ in Eq. \ref{eq:taylorresult} we can write:
\begin{equation}
	T(\theta) = -9.81 \cos(\theta)m
	\label{eq:thrust}
\end{equation}
where again $\theta$ is bounded between $-\pi/2$ and 0.

Even though through the flight control system in practice the produced lift will be close to Eq. \ref{eq:lift}, that does not mean that the control effectiveness of the pitch angle can be obtained from this equation.
This control effectiveness changes strongly with airspeed, and therefore it is estimated experimentally.
Using test flights, we fit the derivative of the angle of attack, measured with an $\alpha$ vane, with the derivative of the measured acceleration at several flight conditions.
The best fit is obtained with an $\alpha$ vane, though it may be possible to estimate the control effectiveness using the pitch angle instead of angle of attack, if such a vane is not available.
The effectiveness at these flight conditions is subsequently approximated with the following function:

\begin{equation}
	\pd{}{\theta}L(\theta,V) =
\begin{cases}
	-24.0 r_\theta m,              & \text{for } V < 12 \mathrm{m/s}\\
	-(V - 8.5)\cdot 6.88 m,                            & \text{for } V \geq 12 \mathrm{m/s}\\
\end{cases}
	\label{eq:liftd}
\end{equation}
where
\begin{equation}
	r_\theta=
\begin{cases}
	0 ,& \text{for } -40 \leq \frac{\theta \cdot 180}{\pi}\\
	(\frac{\theta \cdot 180}{\pi} + 40)/(-40),& \text{for } -80 \leq \frac{\theta \cdot 180}{\pi} \leq -40\\
	1,& \text{for } \frac{\theta \cdot 180}{\pi} \leq -80\\
\end{cases}
\end{equation}

\subsection{Effectiveness of the flaps on lift} \label{sec:flapeff}

\begin{figure}[h]
\centering
\includegraphics[width=0.8\columnwidth]{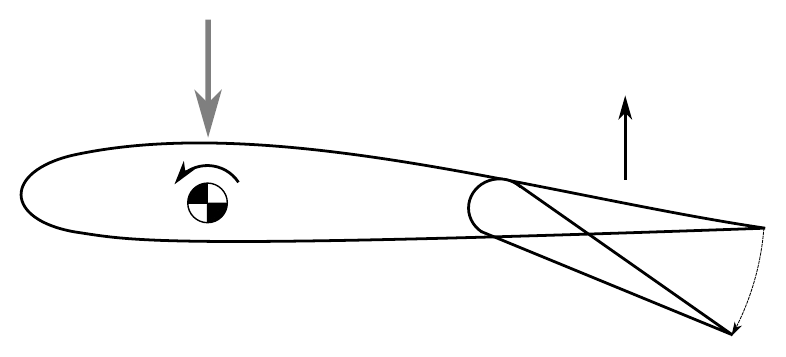}
	\caption{Deflecting a flap downward has an immediate effect (black arrows) and the pitch angle reduction follows later and produces negative lift (gray arrow).}
\label{fig:flap_effect}
\end{figure}

The flaps, whose purpose is to control the rotations around the body Y and Z axes, also have a significant effect on the produced lift.
This situation is depicted schematically in Fig. \ref{fig:flap_effect} and holds for hover as well as forward flight.
In order to achieve a desired acceleration, the vehicle needs to increase or decrease the pitch angle.
The required flap deflections for this change in pitch angle initially lead to an acceleration in the opposite direction, because it increases the lift the flap produces.
In linear time invariant systems theory, this is commonly referred to as undershoot, which is caused by non-minimum phase zeros \citep{goodwin}.
The initially opposite reaction gives rise to oscillations in the desired pitch angle.

\begin{figure}[h]
\centering
\includegraphics[width=1.0\columnwidth]{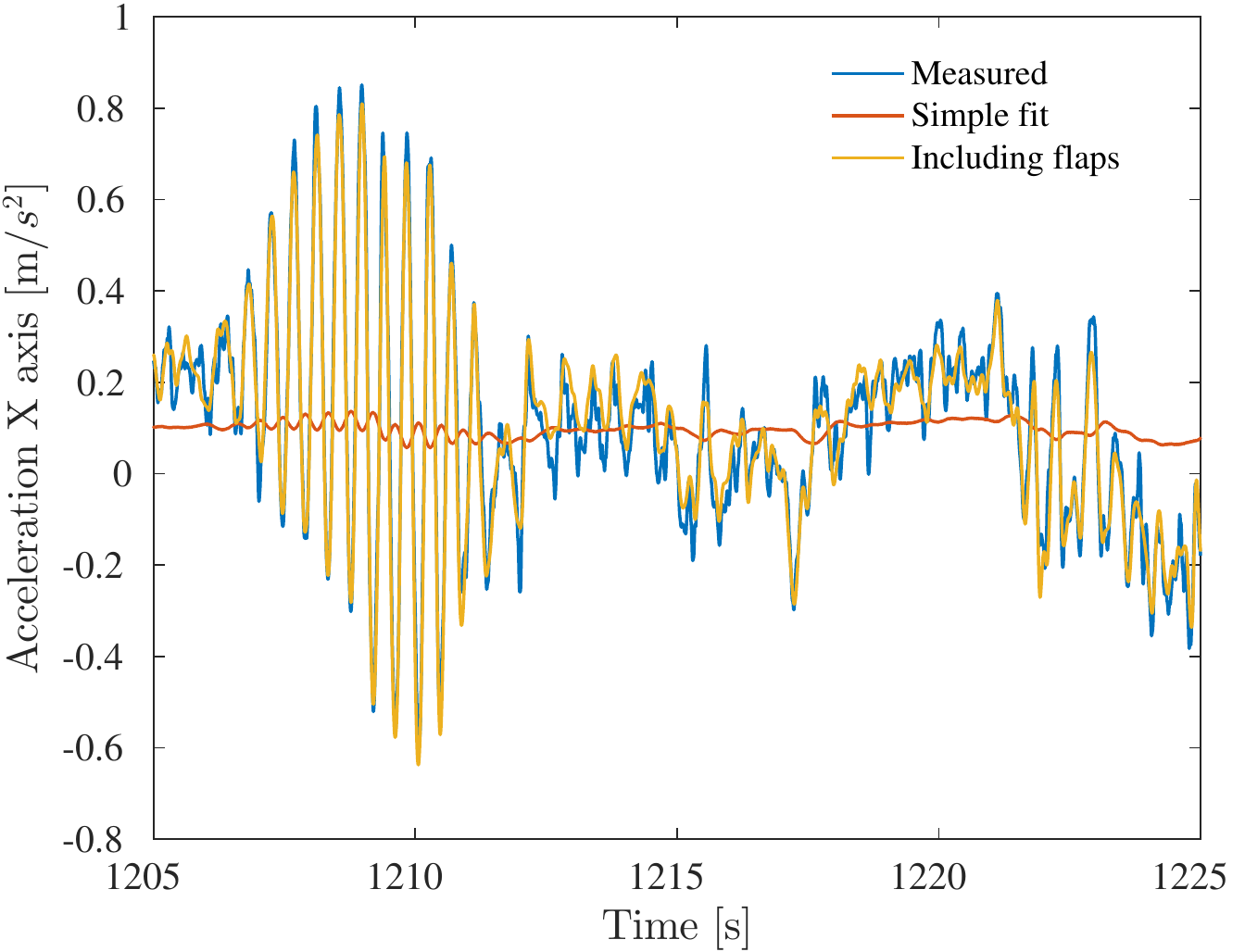}
\caption{Acceleration in the body X axis, along with model fits using input data.}
\label{fig:flap}
\end{figure}

Whenever the controller commands a pitch change in order to change the acceleration, the vehicle at first accelerates in the direction opposite to the desired one.
Consequently, the controller will increase the command, even though the original command would have led to the correct acceleration over time.
This leads to oscillations, both in forward flight and while hovering, which is observed in test flights.
Such an oscillation of the acceleration in body X axis is shown for hover in Fig. \ref{fig:flap}.
Along with the measurement, a simple linear least squares fit is depicted for the acceleration in the body X axis using as inputs an offset, the pitch rate and the pitch, according to the model below:
\begin{equation}
	\ddot{\xi}_{X_B} = [ \begin{array}{lcr} 1 & q & \theta \end{array}]\bs{B}_1
	\label{eq:fit1}
\end{equation}
where $q$ is the pitch rate and $\bs{B}_1$ is a vector of coefficients.
This model is referred to as the simple fit.
As can be seen in Fig. \ref{fig:flap}, these inputs clearly can not explain the measured data, as the fitted data does not coincide with the measured data at all.
That is why the flap deflection has to be added to this model, as is shown below:
\begin{equation}
	\ddot{\xi}_{X_B} = [ \begin{array}{lcccr} 1 & q & \theta & u_{f_0} & u_{f_1} \end{array}]\bs{B}_2
	\label{eq:fit2}
\end{equation}
where $\bs{B}_2$ is a different vector of coefficients and $u_{f_0}$ and $u_{f_1}$ are the filtered inputs to the left and right flaps respectively.
With this model, the fit is much better, as can be observed from Fig. \ref{fig:flap}.
From the accurate model fit we can conclude that the flap deflection indeed plays a large role in the lift production.

To cope with this effect, a possible solution is to increase the control effectiveness of pitch on the acceleration in the controller.
Since the controller gain is the inverse of the control effectiveness, increasing the modeled control effectiveness reduces the initial control effort, such that further increments in pitch angle are needed to achieve the correct acceleration.
Practical experiments showed that by scaling the control effectiveness of the pitch on the acceleration by a factor of two it was possible to remove the oscillation during hover flight.

Although these results are encouraging, the influence of modifying the control effectiveness from its true value on the flight performance is not yet well understood.
Therefore, in this section we present a second solution.
The concept is that if the vehicle does not react to accelerations caused by movement of the flaps, the problem of oscillations is removed.
To achieve ignorance to these accelerations, the acceleration caused by the flaps is modeled, high pass filtered, and subtracted from the acceleration measurement.
The reason to apply the high pass filter is that continuous flap deflections may cause offsets in the compensated acceleration measurement, and only the transient flap movements need to be accounted for.

The compensated acceleration $\ddot{\bs{\xi}}_\text{comp}$ is calculated as follows:
\begin{equation}
\ddot{\bs{\xi}}_\text{comp} = \ddot{\bs{\xi}}_f - \bs{M}_{NB} \left[ \begin{array}{c} HP(-u_{f_0} + u_{f_1})G_{\text{flap}} \\ 0 \\ 0 \end{array} \right]
\end{equation}
where $HP$ is the high pass filter, $u_{f}^0$ is the left flap deflection and $u_{f}^1$ is the right flap deflection, both low pass filtered to synchronize them with the low pass filtered acceleration measurement.
The effectiveness of the sum of flap deflections on the acceleration in the body X axis is denoted by $G_\text{flap}$.
The high pass filter used is a fourth order Butterworth filter with the cutoff frequency tuned to be 0.5 Hz.

Both methods presented in this section succeed in removing the oscillation in test flights.
Arguably, it is simpler to modify the control effectiveness than to implement the compensation for the flap effectiveness.
On the other hand, because of modifying the control effectiveness, it may take longer to counteract disturbances.
For the compensation approach, the control effectiveness is not modified, so the disturbance rejection is expected to remain unchanged.

\subsection{Attitude gains in forward flight}

The roll (around body X axis) and pitch (around body Y axis) are controlled with different actuators with different dynamics, so they allow for different control gains.
The propellers react slower than the flaps, which means that the $K_\eta$ gain should be lower for the roll axis than for the pitch axis. 
The result is that the vehicle is more aggressive in the pitch axis than in the roll axis.

For the hover scenario this is generally not a problem, but for turns in forward flight having different dynamics in pitch and roll can lead to altitude errors.
In a turn, a continuous combination of roll (around the X axis in Fig. \ref{fig:axisdef}) and pitch rate is necessary.
This is realized as the attitude reference rotates ahead of the actual attitude.
The error in roll and pitch angle, multiplied with the $K_\eta$ gain, produces the reference roll and pitch rate.
With a higher gain on the pitch axis, the vehicle pitches up disproportionally.
As the vehicle is pitching up more, the attitude error becomes smaller in the pitch axis and the roll and pitch rates are in proportion again.
The increased pitch angle leads to a higher lift than intended and the vehicle ascends.

\begin{figure}[h]
\centering
\includegraphics[width=0.49\columnwidth]{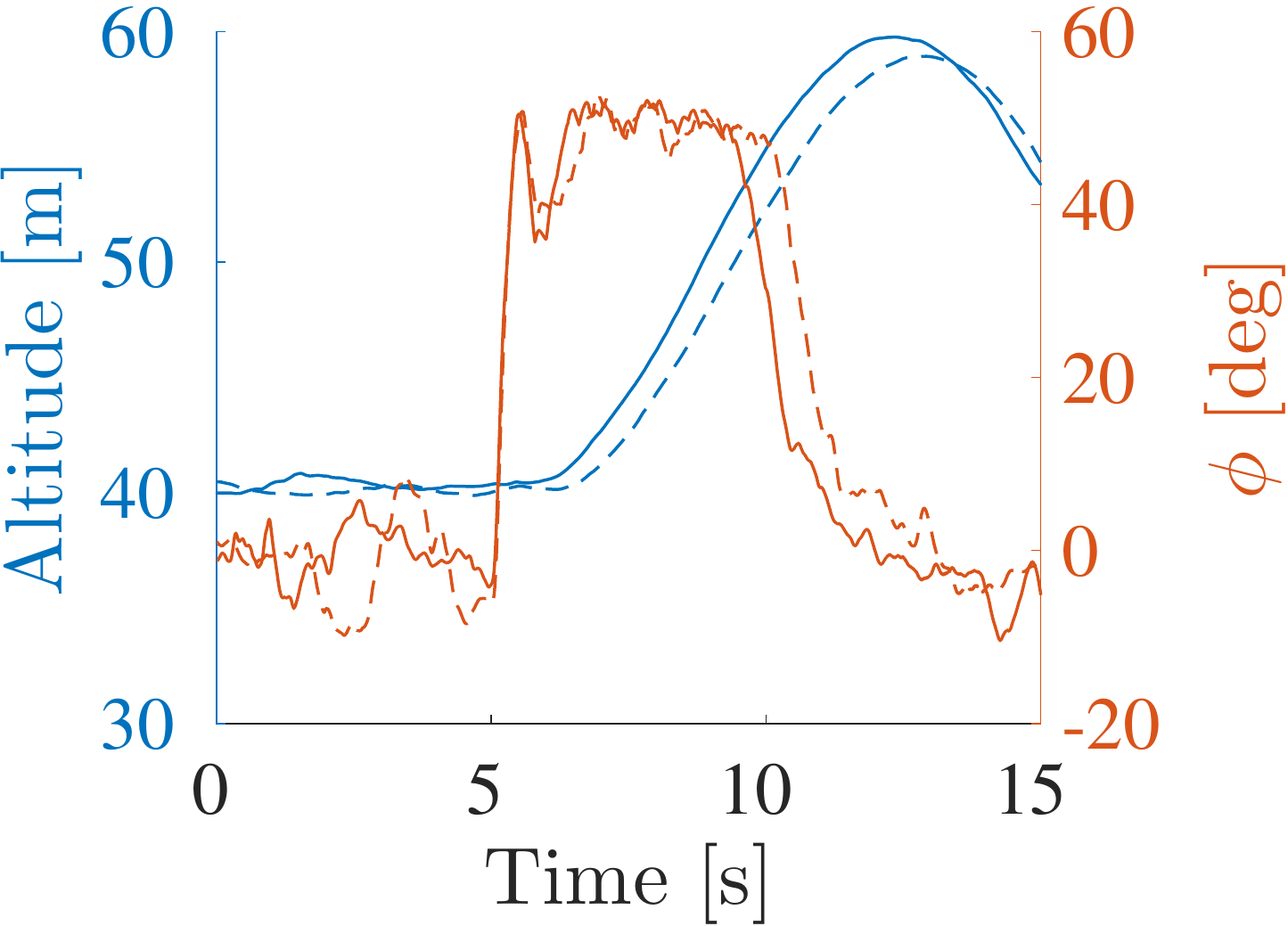}
\includegraphics[width=0.49\columnwidth]{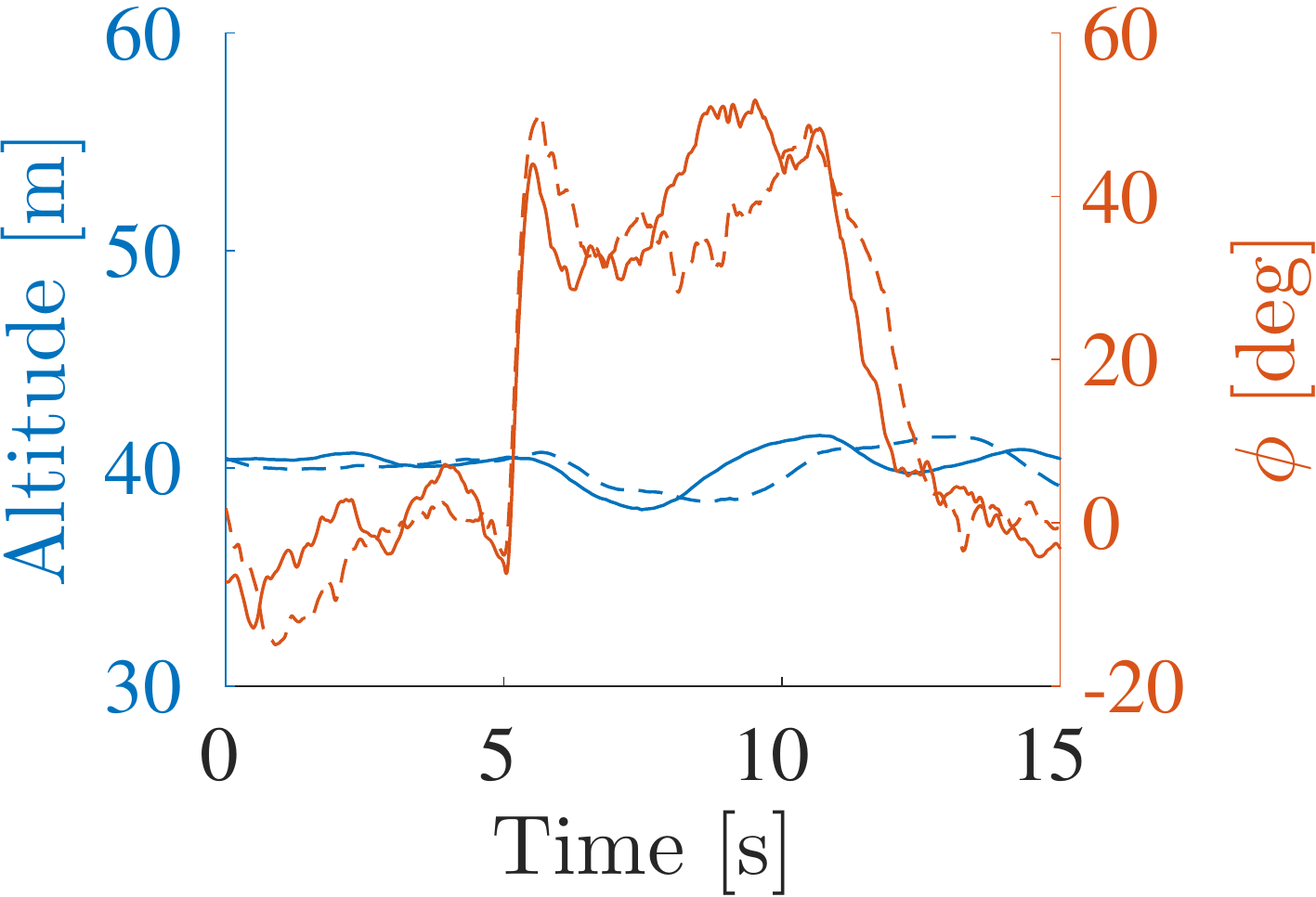}
\caption{Four turns of almost 180$^\circ$ heading change: two with a high pitch gain (left), and two with equal pitch and roll gains (right).}
\label{fig:turn_alt}
\end{figure}

This effect of disproportionate pitch and roll is observed in real flights, especially when making long, almost 180$^\circ$ turns with a pitch gain of 13.3 and a roll gain of 7.6.
During such turns, a pitch angle error is observed, while there is no error in the tracking of angular rates.
For two of these turns, the altitude and roll angle are shown in the left plot of Fig. \ref{fig:turn_alt}.
Most notable is the increase in altitude of more than 18 m above the target altitude of 40 m during the turn.
To remove the asymmetry, both $K_\eta$ gains are given the same value of 7.6 whenever the Cyclone flies faster than 12 m/s.
Two more turns were made with these equal gains, the results of which are shown in the right plot of Fig. \ref{fig:turn_alt}.
Now, the altitude error stays within 2 meters of the desired 40 m altitude.

\section{Guidance} \label{sec:nav}

The guidance of an INDI controlled quadcopter was described in previous research \citep{smeurgust}, and as INDI acceleration control is providing an abstraction layer, the Cyclone can be controlled in the same way.
It starts from a certain position error, which multiplied with a gain produces the desired velocity vector.
Subtract the actual ground velocity, multiplied with a gain, and the desired acceleration is obtained.
Basically, this is a PD controller that provides the acceleration reference ($\ddot{\bs{\xi}}_\mathrm{ref}$) based on the position ($\bs{\xi}$) and velocity ($\dot{\bs{\xi}}$) errors for the INDI controller as is shown below:
\begin{equation}
	\ddot{\bs{\xi}}_{\mathrm{ref}} = ((\bs{\xi}_{\mathrm{ref}} - \bs{\xi})K_\xi - \dot{\bs{\xi}}) K_{\dot{\xi}}
	\label{eq:nuxi}
\end{equation}

\subsection{Efficient turning}

During initial test flights, Eq. \ref{eq:nuxi} was the method of guidance for the Cyclone.
Though this method is feasible, there is a specific downside to the approach.
This form of guidance will always result in the shortest ground track, which is not necessarily the most efficient for a hybrid vehicle.
Take the example of the vehicle cruising at 20 m/s with a certain heading.
If the vehicle is now commanded to fly in the opposite direction, it will have to break all the way to 0 m/s, and then accelerate back to 20 m/s in the opposite direction.
Since the Cyclone is less efficient while hovering, this approach is expected to be less efficient than a turn at the same airspeed.

To improve the efficiency of the guidance, a rule based strategy is employed, depending on the current airspeed and the desired airspeed.
If the current airspeed is higher than 10 m/s and the desired airspeed is higher than 14 m/s, the vehicle will make a turn (fixed wing style).
In this case, the airspeed is controlled, also during the turn, and if applicable the drone will accelerate or decelerate during the turn.
The reason that the desired airspeed has to be larger than 14 m/s is that above this airspeed, the measurement of the airspeed is considered reliable, avoiding any kind of switching behaviour.
In all other cases, the vehicle will take the direct approach, as is given by Eq. \ref{eq:nuxi}.

\subsection{Approaching a waypoint}
The desired velocity towards a goal position is calculated by a proportional controller.
The proportional controller works well if the velocity towards the waypoint is low, but if it is large, it may lead to overshoot.
The reason is that the proportional gain will require a certain deceleration per meter, which means that if the speed towards the waypoint is high, the required deceleration is high.
However, the maximum deceleration of the Cyclone is limited, hence the overshoot.

If we assume this maximum deceleration to be constant over the flight envelope, we can calculate the maximum allowable speed as a function of the distance at which the deceleration is started.
Using classical mechanics, we arrive at $v=\sqrt{2da_{\text{max}}}$, where $d$ is the distance to the waypoint and $a_{\text{max}}$ is the maximum deceleration.
If the speed commanded by the proportional controller is higher than this maximum allowable speed, it is reduced to the maximum allowable speed.
Limiting the speed in this way avoids overshoot and limits pitch-up problems during the transition to hover.
Specifically, it enables the vehicle to approach a waypoint with tailwind, without any overshoot.

\subsection{Line following}

In order to give some kind of guarantee on the path that the vehicle will follow, a path tracking algorithm is necessary.
Eventually the goal is to make the Cyclone track any path that is within the performance limits of the vehicle, possibly using a method from literature \citep{marina}.
For now, the Cyclone has straightforward line following functionality, as most paths can be approximated with a combination of lines.

\begin{figure}[h]
\centering
\includegraphics[width=1.0\columnwidth]{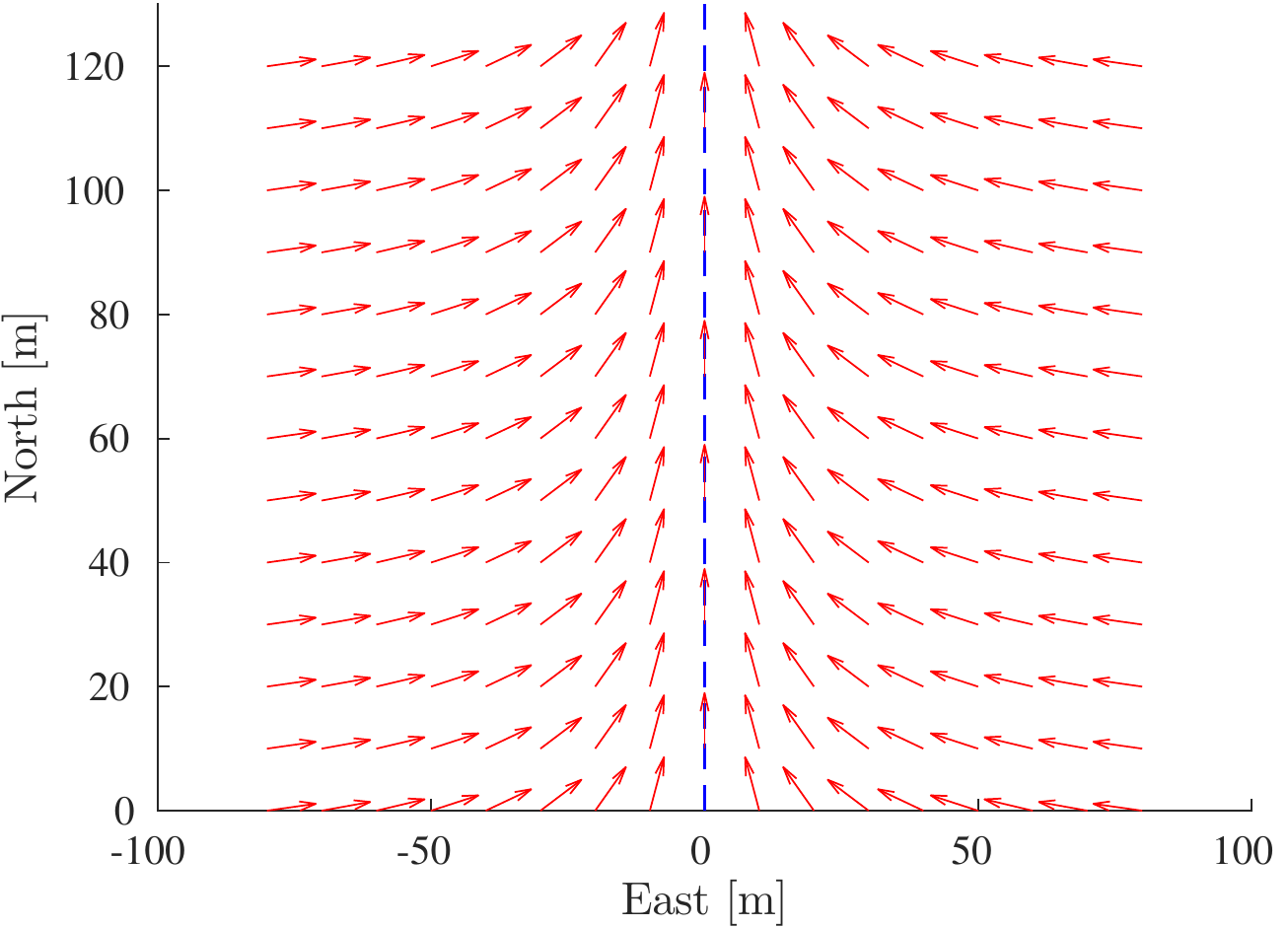}
\caption{Velocity vector field corresponding to an arbitrary line.}
\label{fig:vf}
\end{figure}

Each line is defined with a start and end point, and a corresponding field of ground velocity vectors is calculated that converges to and along the line segment.
The angle $\lambda$ of these vectors with respect to the line is given by:
\begin{equation}
	\lambda = \text{atan} \left( \frac{d+0.05d^2}{50} \right)
\end{equation}
where $d$ is the absolute distance orthogonal to the line.
The resultant vector field for a fixed speed is shown in Fig. \ref{fig:vf}.

The magnitudes of the velocity vectors can be predefined or they can be proportional to the distance to the end point.
By setting their magnitude to a relatively large number, while limiting the maximum airspeed, the Cyclone can be made to fly constantly at the defined maximum airspeed.
The line following is only valid if the normal line that goes through the end point is not crossed and the vehicle is a minimum distance away from the end point.
At this distance, the vehicle switches to the next element in the flight plan, which could be another line.

\section{Test flight results} \label{sec:results}

During the development of the controller discussed in this paper, many test flights have been performed.
This section will present some of these test flights to support claims made throughout this paper.

\begin{figure*}[h!]
\centering
\includegraphics[width=1.0\textwidth,trim={2cm 5cm 0cm 5cm},clip]{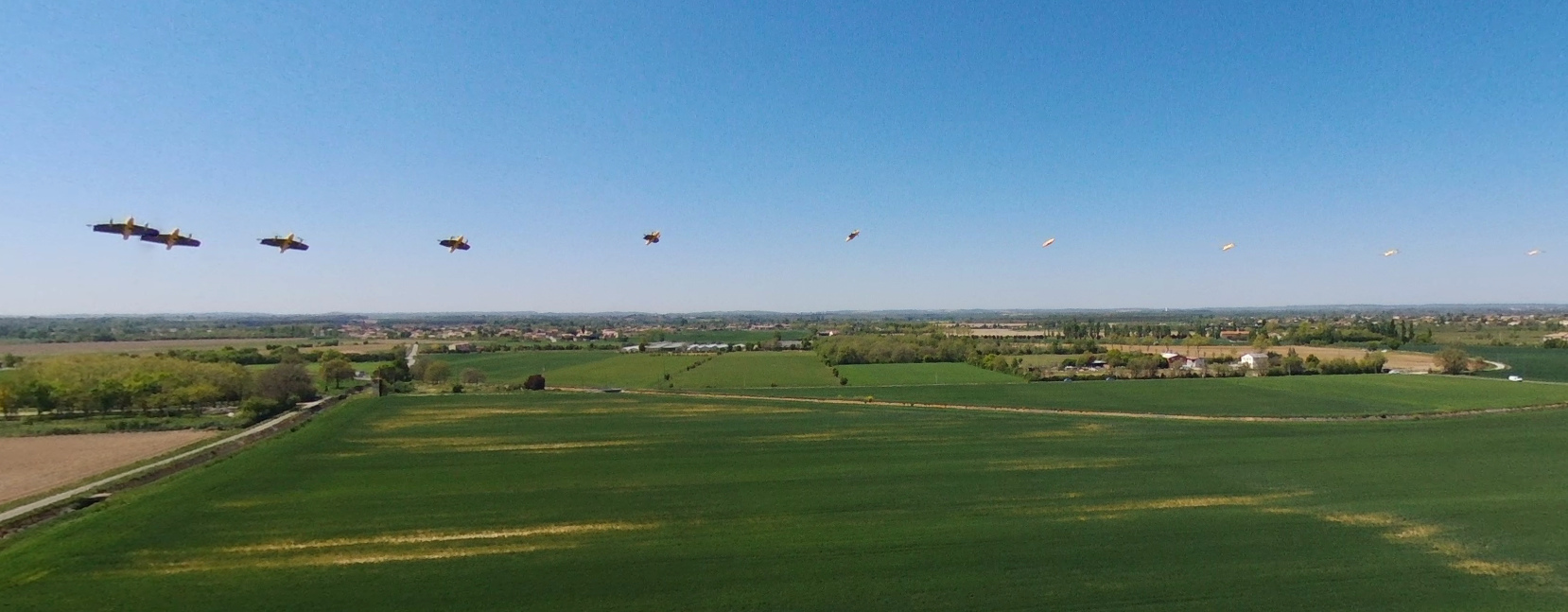}
\caption{Composite image of a transition to forward flight with a constant acceleration of 1 m/s$^2$, with 1 s image intervals.}
\label{fig:transpic}
\end{figure*}

\begin{figure*}[h!]
\centering
\includegraphics[width=0.8\textwidth,trim={8cm 5cm 0cm 3cm},clip]{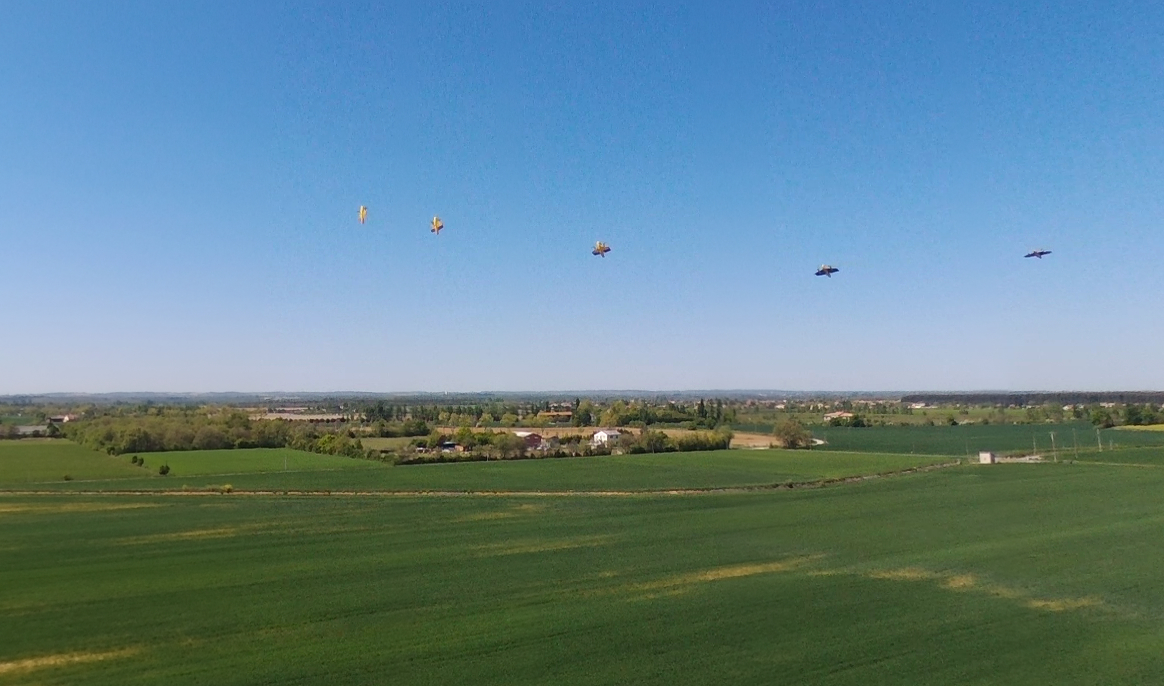}
\caption{Composite image of a transition to hover flight, with 1 s image intervals.}
\label{fig:transpicback}
\end{figure*}

Figure \ref{fig:transpic} shows a picture of a transition to forward flight, where the Cyclone was specifically commanded to constantly accelerate at 1 m/s$^2$.
The vehicle starts out hovering in the left side of the picture and transitions to forward flight as it flies to the right side of the picture.
The picture is constructed from video frames, taken by a hovering Bebop quadrotor, that rotated in order to keep the Cyclone in the frame.
The frames are taken with 1 second intervals and stitched together afterwards.
The figure also gives a visual impression of the Cyclone's ability to transition while aligning itself with the direction of movement.
A transition back to hover is shown in Fig. \ref{fig:transpicback}, again with 1 second intervals.
Here, the vehicle comes flying into the frame in forward flight from the right side, and transitions back to hover as it flies to the left.

\subsection{Attitude control performance}
First, the attitude control is shown for the case where the Cyclone is flying back and forth between two waypoints, utilizing wing borne flight in between at hovering at each waypoint.
The waypoints are more than 200 m apart in the east direction, and the west waypoint is 17 m more northern.

\begin{figure}[h!]
\centering
\includegraphics[width=1.0\columnwidth]{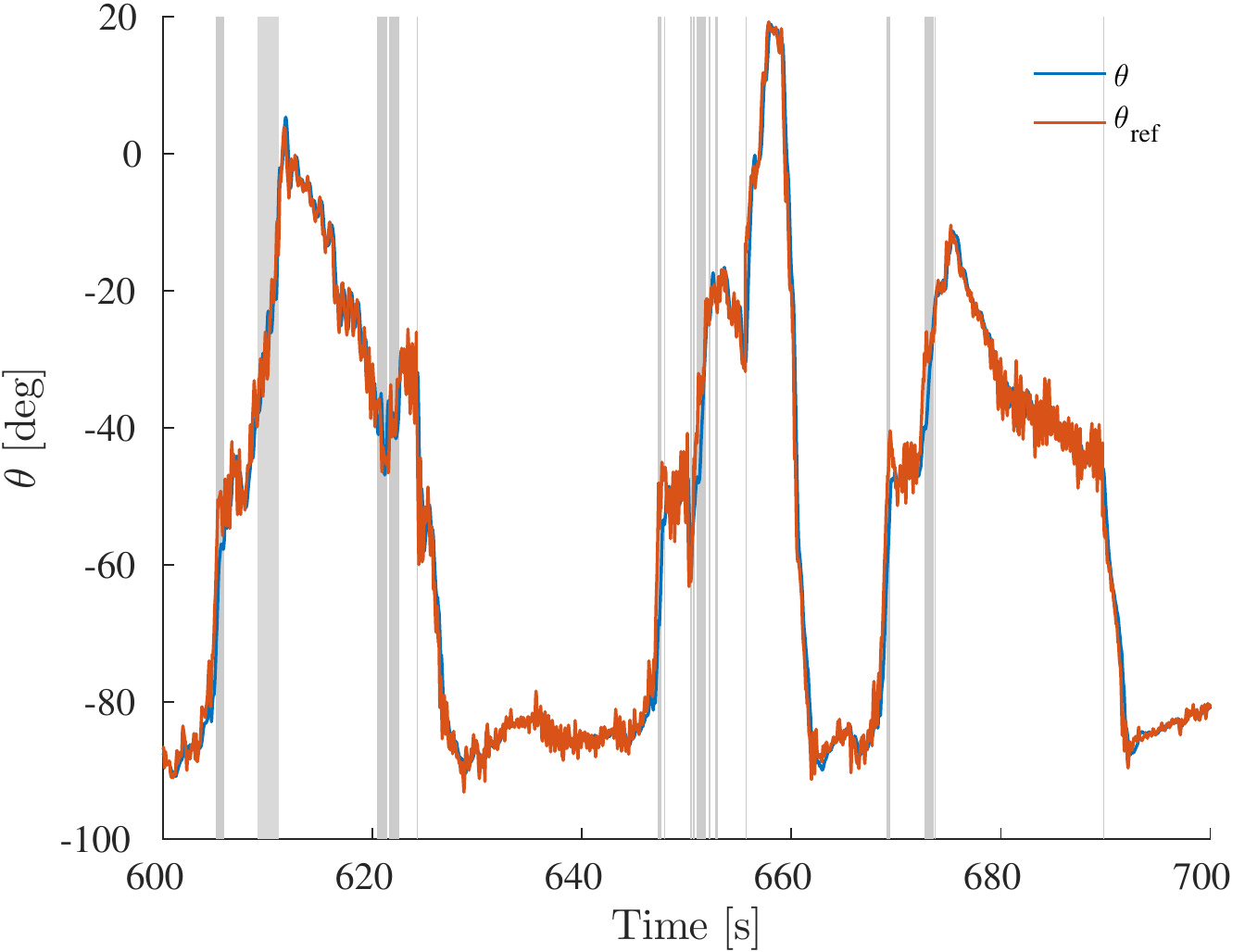}
\caption{Pitch angle for the experiment (ZXY Euler), where shaded areas indicate utilization of thrust in order to pitch.}
\label{fig:theta}
\end{figure}

\begin{figure}[h!]
\centering
\includegraphics[width=1.0\columnwidth]{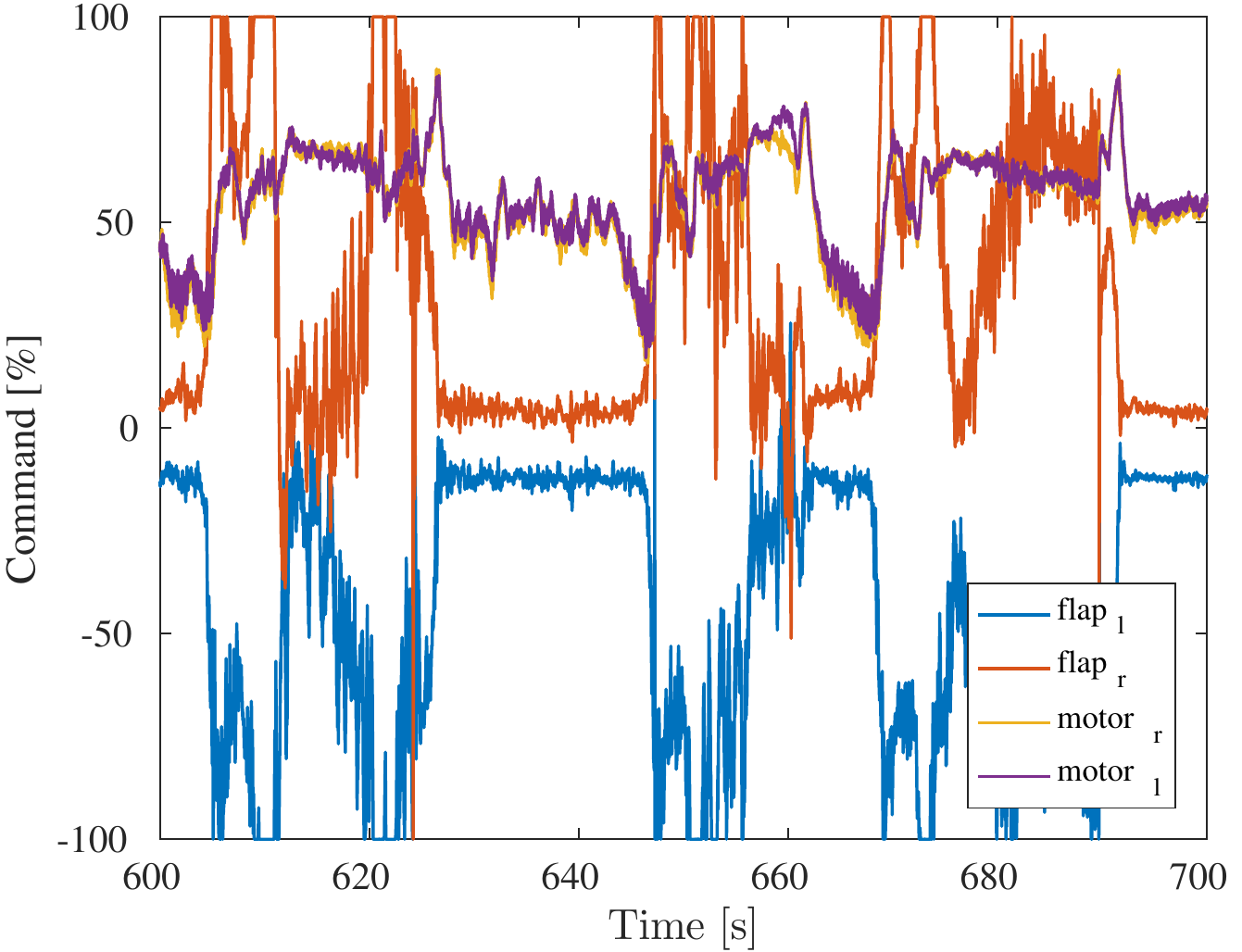}
\caption{Inputs to flaps and motors during the experiment, where left flap down is positive and right flap up is positive.}
\label{fig:inputs}
\end{figure}

From Fig. \ref{fig:theta}, it can be seen that the pitch angle can be tracked across the flight envelope.
Figure \ref{fig:inputs} shows the actuator inputs that were given during the flight.
Even though saturation of the flaps can be observed whenever the aircraft is pitching up at high angle of attack, the pitch angle remains tracked.
The shaded areas in Fig. \ref{fig:theta} represent times at which the controller used thrust in order to provide extra pitch moment.
The use of thrust to pitch typically happens at the intermediate pitch angles.

\begin{figure}[h!]
\centering
\includegraphics[width=1.0\columnwidth]{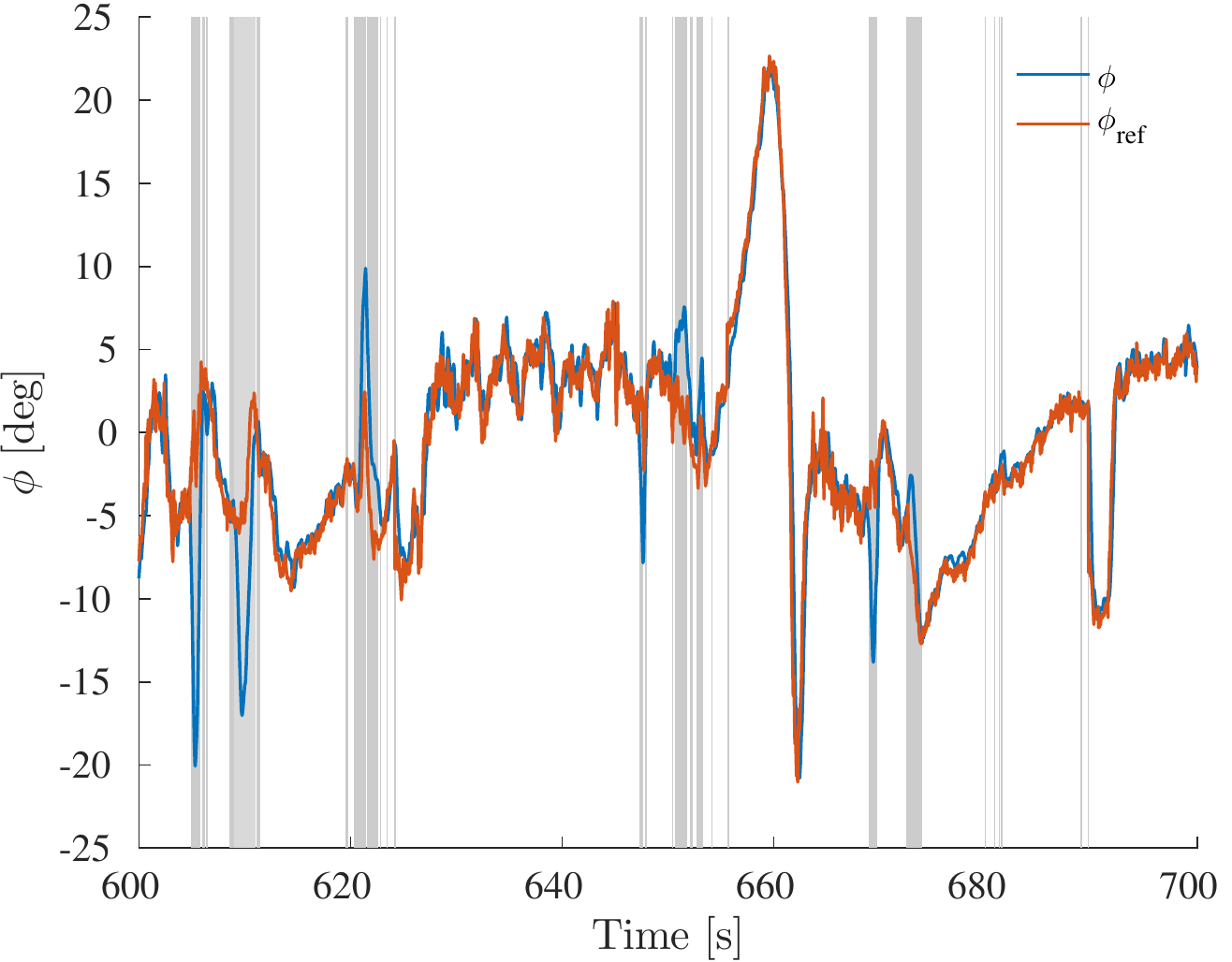}
\caption{Roll angle for the experiment (ZXY Euler), where shaded areas indicate saturation of at least one of the flaps.}
\label{fig:phi}
\end{figure}

Figure \ref{fig:phi} shows the roll angle (ZXY Euler convention) for the same flight.
When saturation of the flaps occurs, the roll is not well controlled.
The reason is that the pitch axis has priority over the roll axis in the case of saturation.
Though there is nonzero roll error occasionally during the flight, it does not lead to instability.

\begin{figure}[h!]
\centering
\includegraphics[width=1.0\columnwidth]{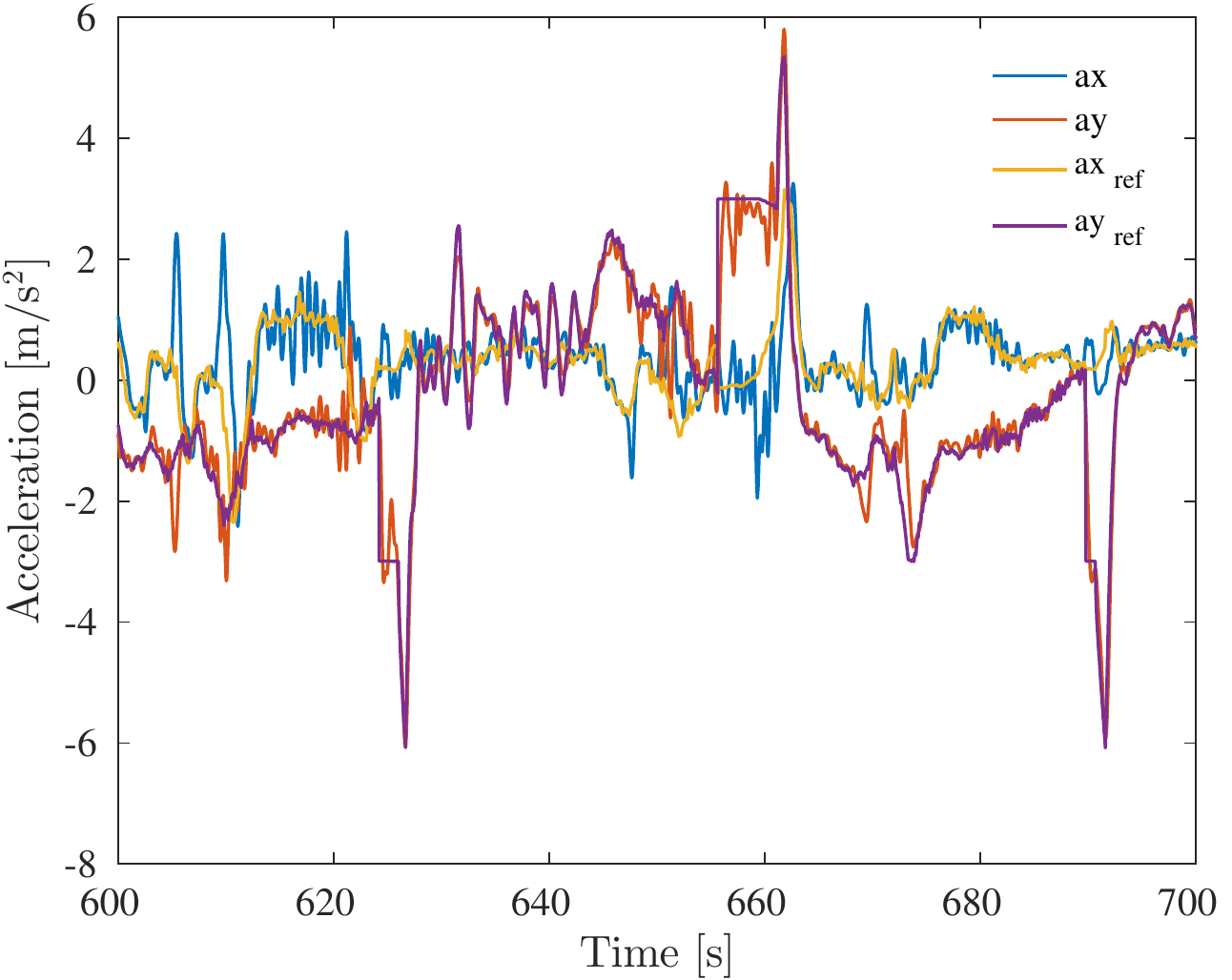}
\caption{Acceleration in the North (X) and East (Y) axes.}
\label{fig:accelxy}
\end{figure}

Section \ref{sec:actrl} described how the Cyclone controls accelerations.
Here, it is investigated how well the accelerations are tracked for the flight between waypoints described above.
Figure \ref{fig:accelxy} shows the accelerations in the North and East axes, along with the reference acceleration that should be tracked.
From this figure, it becomes clear that most of the time the accelerations are well tracked.
At some instances, the acceleration is not tracked well, such as at t=606 and t=610.
The temporary loss of tracking can be attributed to saturation of the flaps, which prohibits complete realization of the control objective.
Temporary loss of tracking is also observed for the roll angle when the flaps saturate in Fig. \ref{fig:phi}, but tracking is restored once the actuators are not saturated any more.

\begin{figure}[h!]
\centering
\includegraphics[width=1.0\columnwidth]{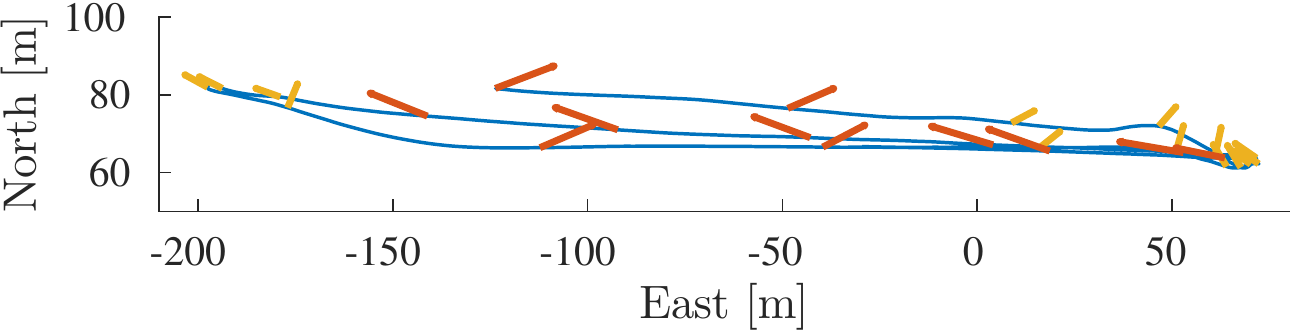}
	\caption{Top view of the experiment, with red arrows for measured airspeed (assuming no sideslip) and orange arrows for estimated airspeed at low speeds.}
\label{fig:topview}
\end{figure}

A top view of the flight is shown in Fig. \ref{fig:topview}.
It shows the track, along with the airspeed vector every four seconds in red.
For parts of the flight where the airspeed is lower than 10 m/s, the airspeed measurement is not accurate.
For these vectors, it is assumed that the airspeed was as large as the average estimated wind speed of 6.7 m/s, and these vectors are displayed in orange.
The wind came from approximately -70 degrees north.
When going to the eastern waypoint, the vehicle rotates more than 120 degrees while departing for the other waypoint.
In these flights, the Cyclone did not have a certain path defined, which is why not all routes between the two waypoints exactly coincide.

\subsection{Forward flight}

\begin{figure}[h!]
	\centering
	 \begin{overpic}[width=1.0\columnwidth]{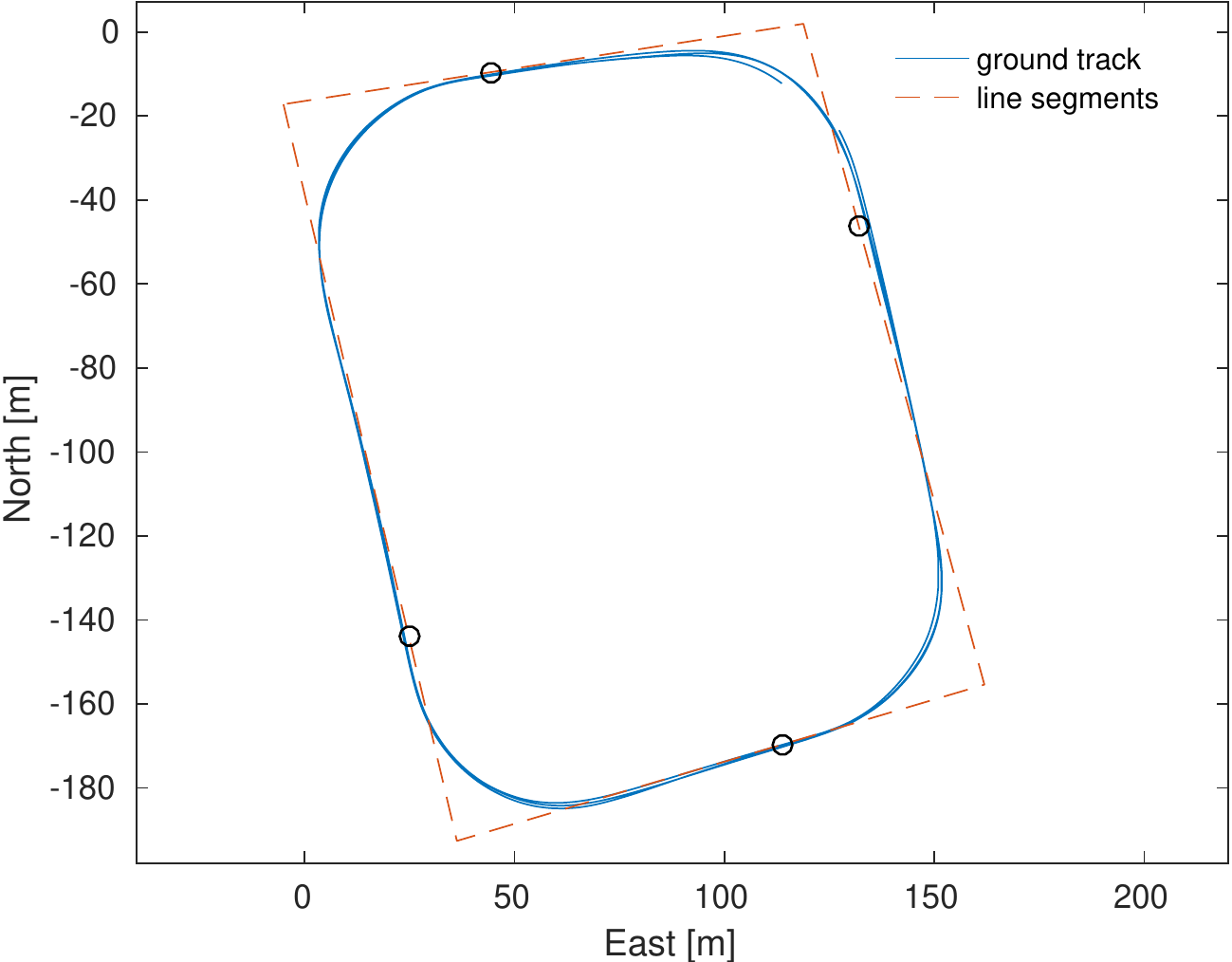}
		 \put(88,52){\includegraphics[scale=0.5]{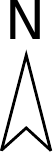}}
	\end{overpic}
	\caption{Top view of line following without wind, where the line switch distance is indicated with small circles (the aircraft flew counter-clockwise).}
\label{fig:topfwd_clean}
\end{figure}

To demonstrate the forward flight capabilities, Fig. \ref{fig:topfwd_clean} shows a ground track where the vehicle consecutively follows lines, such that it tracks a polygon.
The maximum airspeed was set to 16 m/s, while the desired speed along the line was 26 m/s.
In the figure, the small circles represent the points at which the vehicle switches to tracking the next line.
The figure demonstrates the ability to converge to and accurately track line segments.
This flight was performed on a day with hardly any wind.

\begin{figure}[h!]
	\centering
	 \begin{overpic}[width=1.0\columnwidth]{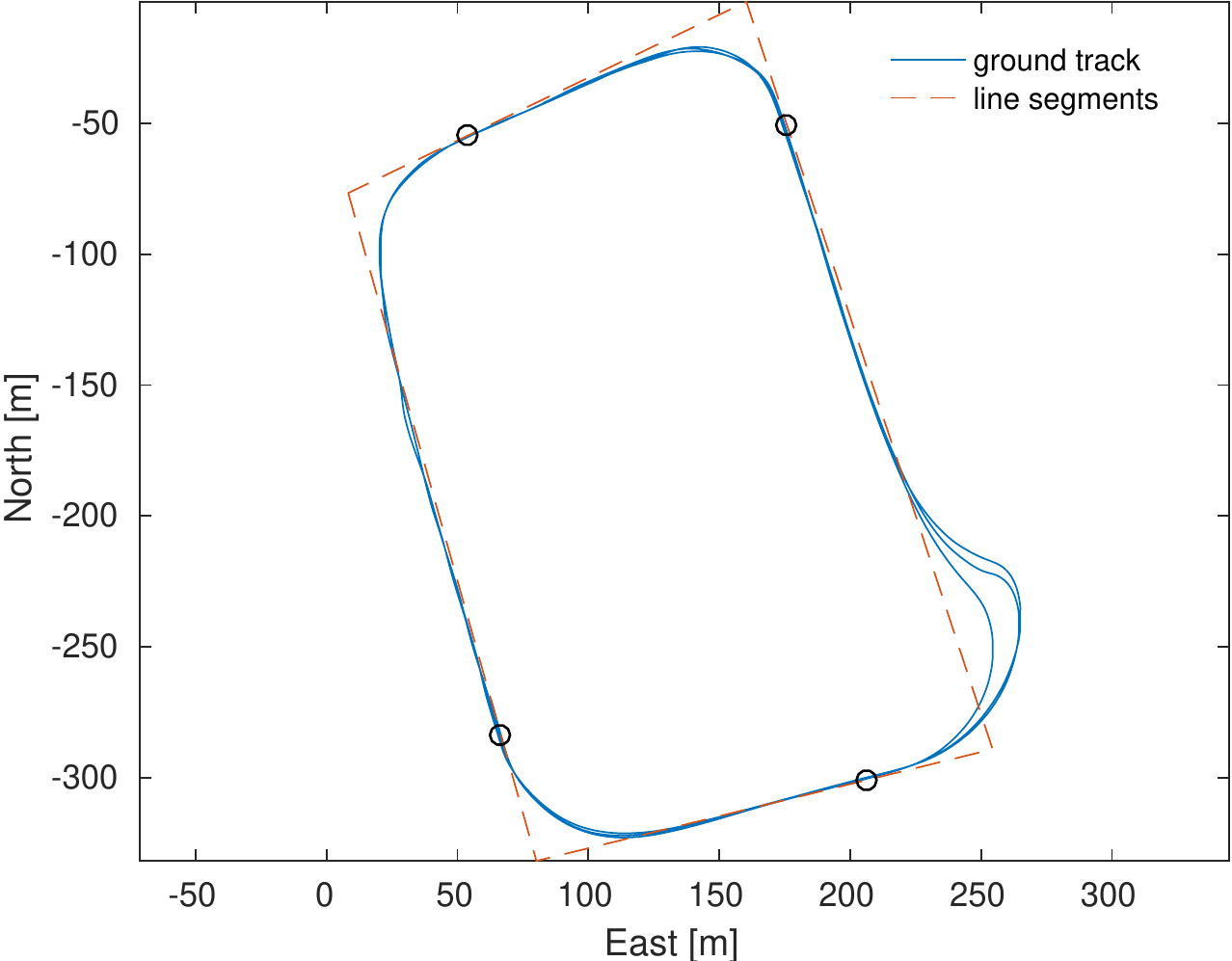}
		 \put(88,52){\includegraphics[scale=0.5]{north.pdf}}
	\end{overpic}
	\caption{Top view of line following, with considerable wind from the west, where the line switch distance is indicated with small circles (the aircraft flew counter-clockwise).}
\label{fig:topfwd}
\end{figure}

Figure \ref{fig:topfwd} shows the ground track for a similar polygon on a windy day.
The average wind speed was estimated at 8.3 m/s from west-southwest direction, based on the airspeed readings during the flight.
The figure shows that regardless of the wind, the lines can be tracked accurately.
However, the figure also shows the limitations of this straightforward method.
The turn in the southeast corner structurally overshoots the line, which can be understood by noting that because of the wind, the ground speed is very high at this point.
The tight turn requires accelerations that are simply not feasible, which shows the need for a proper path following algorithm that will always respect the acceleration limitations of the vehicle.


\section{Guidelines for implementing INDI for hybrids} \label{sec:impl}

The proposed approach to attitude control, velocity control, and guidance is applicable to a wide range of hybrid UAVs.
Some UAV designs may require some adaptations to the controller, but in general the implementation approach is the same.
In order to facilitate the implementation on other vehicles, here, we summarize the basic steps that are needed to start flying a hybrid UAV with the INDI controller.
It is assumed that it is already possible to fly the vehicle vertically, possibly by performing safe tests where the drone is attached to a rope.
Relating to the sections of this paper, the general steps can be listed as follows:
\begin{enumerate}
	\item Identify actuator dynamics, by measuring the response over time to step inputs. For motors, measure the RPM as a function of time. For servos, the position can be obtained as a voltage from the internal potentiometer.
	\item Choose a filter cutoff for the gyroscope and accelerometer noise level in flight. More filtering means that less noise is propagated to the actuators, but it also means that the system reacts slower to disturbances.
	\item Identify the control effectiveness of the actuators using test flights. This can be done by fitting changes in measured data, i.e. (rotational) acceleration, with a function of changes in the control inputs and important state variables, such as airspeed.
	\item Design the control gains $K_\Omega$ and $K_\eta$ (Section \ref{sec:indi}) such that the attitude response is fast and stable, and repeat the previous step for forward flight.
	\item Add thrust effectiveness on pitch when the flaps are saturated (if needed), by estimating the effectiveness of the thrust on the angular acceleration in the pitch axis for large flap deflections from flight data.
	\item Identify the effectiveness of changes in pitch angle on the vertical acceleration. Repeat this step for a number of airspeeds in the flight envelope, such that you can estimate a function like Eq. \ref{eq:liftd}.
	\item Add effectiveness of flap deflection on the lift production, as explained in Section \ref{sec:flapeff}.
	\item Test INDI acceleration control and subsequently fully autonomous flight.
\end{enumerate}

\begin{figure}[h!]
\centering
\includegraphics[width=0.8\columnwidth]{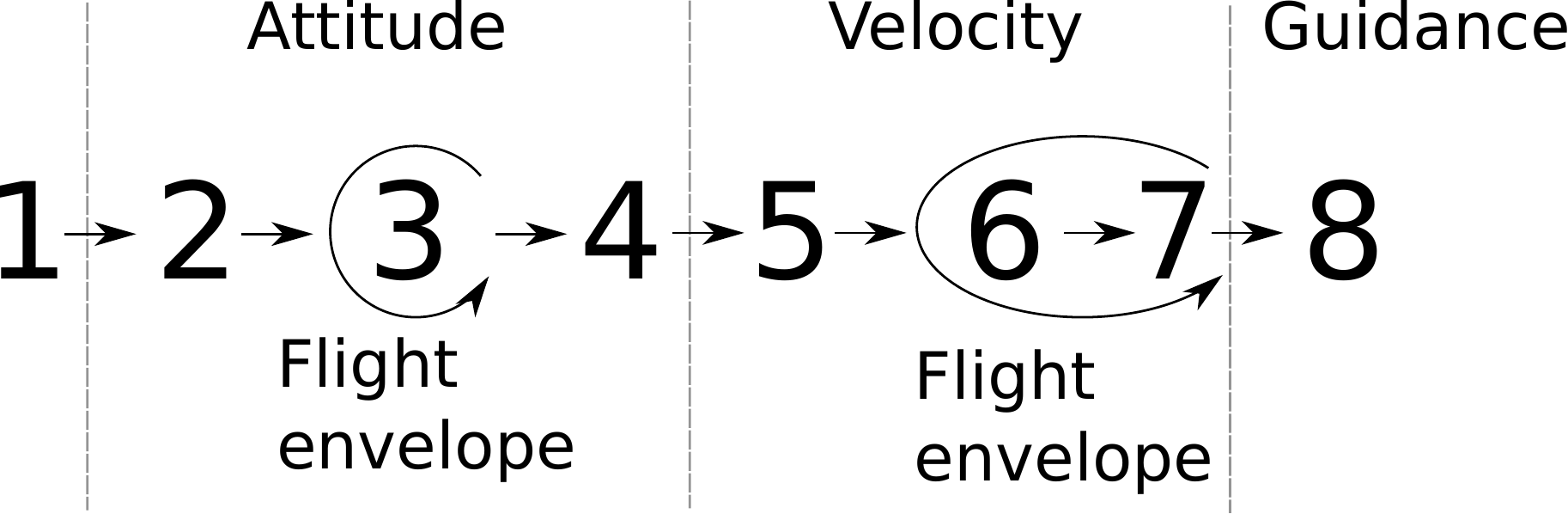}
\caption{Schematic overview of the steps that need to be taken to apply INDI to a new hybrid vehicle.}
\label{fig:guidelines}
\end{figure}

Steps three, six and seven should be repeated for several points in the flight envelope, which is indicated schematically in Fig. \ref{fig:guidelines}.
The figure also indicates which steps are part of the attitude control, position control and autonomous flight.

We have argued that INDI is an approach that does not require a lot of modelling, but still there are a few parameter estimation steps in the list above.
One may wonder if it is truly less involved than a model based approach.

In the list above, note that the only thing we are estimating are control derivatives and actuator dynamics.
To make a full model of the vehicle dynamics, next to the control derivatives, we would need to know the lift, drag, and moments as a function of the vehicle state and actuator inputs.
This would be more difficult than the simple functions for the control effectiveness we have used in this paper.
Further, the vehicle state vector necessary to make such a model accurate, must contain information about the aerodynamics, at least the angle of attack and the airspeed.
These things are hard to measure at low airspeed, making it difficult to apply the model in real life.

\section{Efficiency} \label{sec:efficiency}

\begin{figure}[h!]
\centering
\includegraphics[width=0.8\columnwidth]{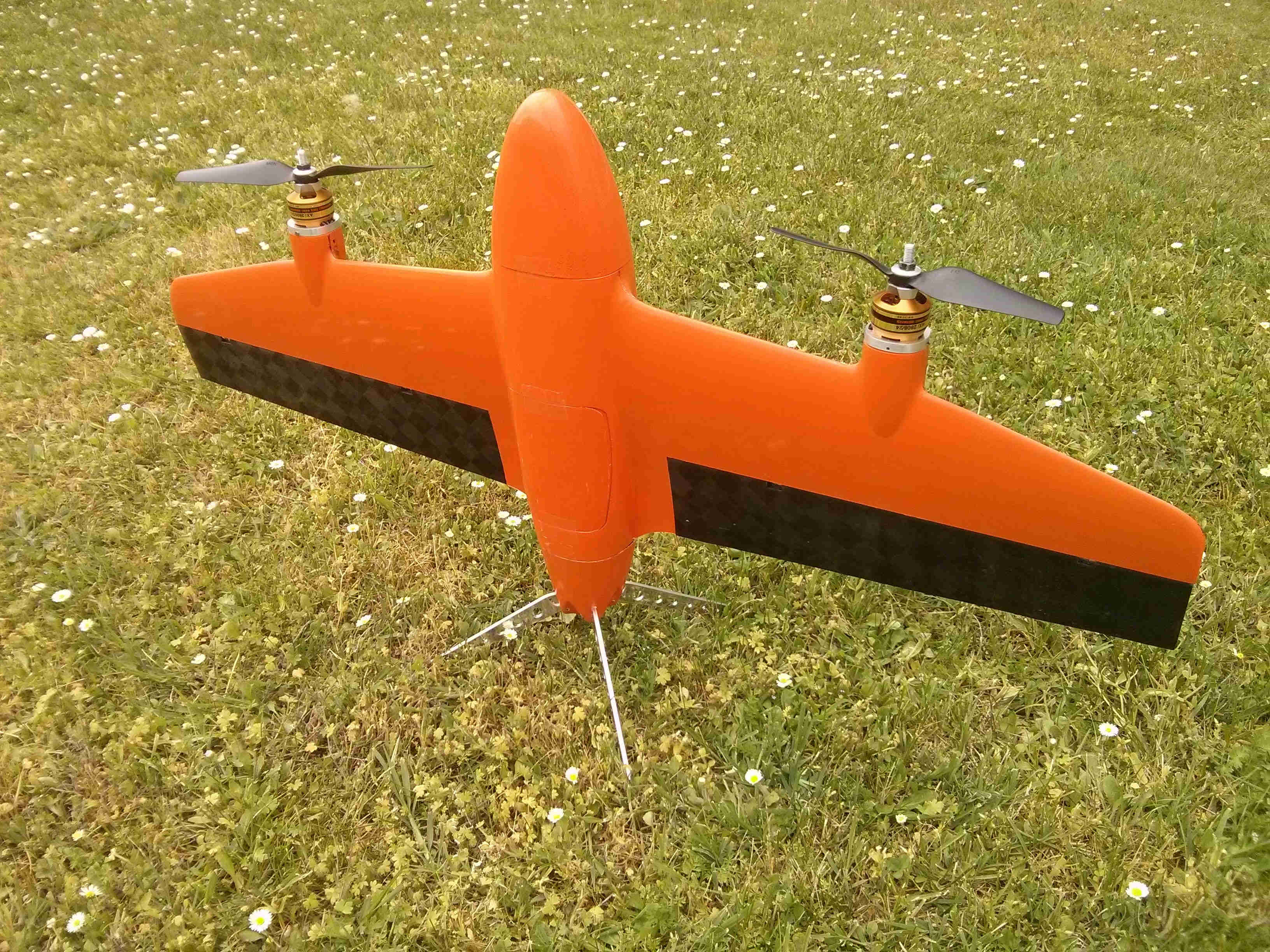}
\caption{A high quality build of the Cyclone.}
\label{fig:cyclone}
\end{figure}

The Cyclone that is used for the test flights, as depicted in Fig. \ref{fig:hover} and Fig. \ref{fig:cyclgrass}, is constructed with a crash-resistant design.
To verify that the designed efficiency specifications are met, a high quality vehicle is constructed as well, shown in Fig. \ref{fig:cyclone}.
The skin has a sandwich structure of aramid and glass fiber, cured in CNC machined aluminum molds, and the total mass of the vehicle is about 1.2 kg.
More information on the construction process can be found in our previous publication \cite{bronz2017}.

\begin{figure}[h!]
\centering
\includegraphics[width=1.0\columnwidth]{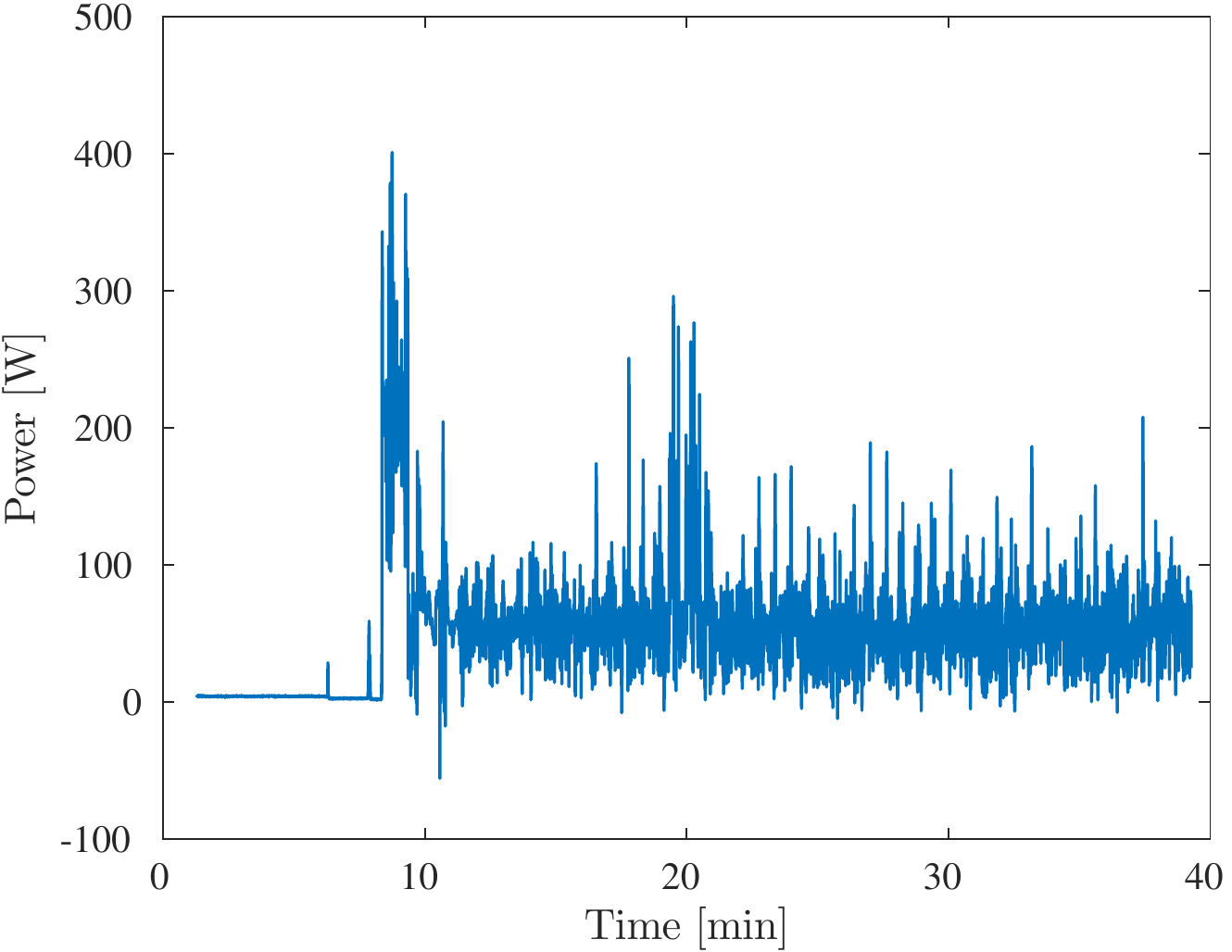}
\caption{The power used during one of the flights of the Cyclone.}
\label{fig:efficiency}
\end{figure}

The vehicle of Fig. \ref{fig:cyclone} has flown several times, of which once with a current sensor and voltage sensor on board.
The power usage can be calculated by multiplying the current and voltage at each time instance, as is shown in Fig. \ref{fig:efficiency}.
In this flight, the vehicle takes off vertically, hovers, transitions to forward flight, and then flies a circular pattern for 29 minutes.
During the flight, the desired airspeed was adjusted to investigate the effect on the efficiency, leading to stall around the twenty minute mark and a subsequent increase in required power.
In the takeoff and hover part of the flight, the vehicle uses an average of 205 Watt.
In the forward flight part of the flight, the average power usage is 55 Watt.
The important thing to note is that the vehicle is intended to spend most of its time in forward flight, such that long flight times can be achieved and large distances can be traveled.

\section{Conclusion} \label{sec:concl}

We have described attitude and position control using Incremental Nonlinear Dynamic Inversion (INDI) for a hybrid tailsitter Unmanned Air Vehicle (UAV).
Though the control derivatives of the actuators and the actuator dynamics need to be known, the INDI controller does not require modeling of the complex aerodynamics of the tailsitter aircraft.
Test flights show that unmodeled forces and moments, such as the strong pitch-down moment during transitions, are effectively counteracted by the incremental control structure.
The resultant controller is able to track a three dimensional acceleration reference, and in doing so can autonomously transition to forward flight and back to hover.
The control effectiveness is adjusted in flight to cope with the changing flight conditions, but the attitude and velocity control structure always remain the same.
Only crude modeling of the control effectiveness is needed, as the incremental, sensor based approach of INDI can correct for many of the modeling errors.

A sideslip controller was designed that ensures that the vehicle always aligns its wing with the airspeed vector, across the entire flight envelope including hover.
Though the transitions between flight phases are not explicitly defined, the vehicle can naturally perform these maneuvers, because of the combination of INDI acceleration tracking control and the sideslip controller.
This removes the strict need to take flight phases into account in the guidance, making the design and execution of flight plans easy and intuitive.

\section*{Acknowledgements}
We would like to thank Xavier Paris, Hector Garcia de Marina and Kevin van Hecke for their help with the test flights.
This research was funded by the Delphi consortium.

\FloatBarrier

\bibliography{bibliography}

\end{document}

%% file: aoa.tex
\begin{tikzpicture}[scale=1.1]




\node[inner sep=0pt] (cyclone) at (-1.8,-0.8)
    {\includegraphics[width=.25\textwidth]{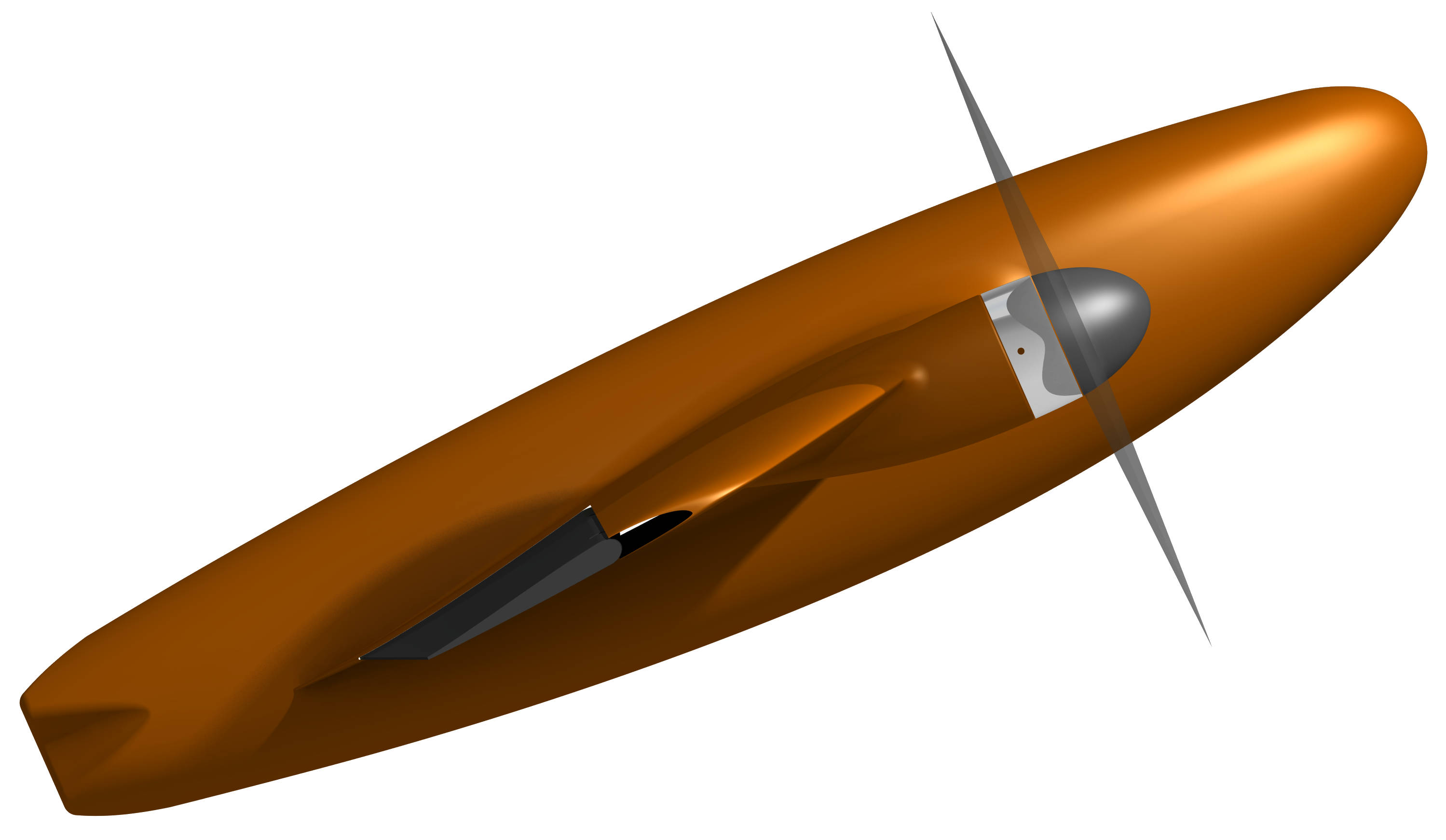}};

\coordinate (O) at (0,0) ;
\coordinate (O1) at (3,0) ;
\coordinate (O2) at (0,-2.0) ;
\coordinate (O3) at (0.35,0) ;
\coordinate (O4) at (0,-0.5) ;
\coordinate (A) at (8:3.4) ;
\coordinate (B) at (16:3.1) ;

	\coordinate (xb1) at (24.5:3) ;
	\coordinate (v1) at (12:3) ;
	\coordinate (h1) at (0:3) ;

\draw[->] (O4) -- (O2) node[right] {$Z_N$};
\draw[->] (30:0.35) -- (24.5:3) node[above] {$-Z_B$};
	\draw[>=stealth,->, thick] (12:0.35) -- (A) node[draw=none,fill=none,right,below] {$\bs{V}_g$} ;
	\draw[>=stealth,->, thick] (20:0.35) -- (B) node[draw=none,fill=none,right,above] {$\bs{V}$};
\draw[>=stealth,->, thick] (B)      -- (A) node[draw=none,fill=none,midway,right] {$\textbf{w}$};


\draw[->] (2.4,0) arc (0:8:2.4);
\node[] at (4:2.6)  {$\gamma$};

\draw[->] ++ (16:2.4) arc (16:24.5:2.45);
\node at (20:2.7)  {$\alpha$};

	\draw (O3) -- (O1) node[below] {horizontal} ;

\end{tikzpicture}

%% file: knife.tex
\begin{tikzpicture}

\node[inner sep=0pt] (cyclone) at (0.6,-1.1)
		{\scalebox{-1}[1]{\includegraphics[width=.25\textwidth]{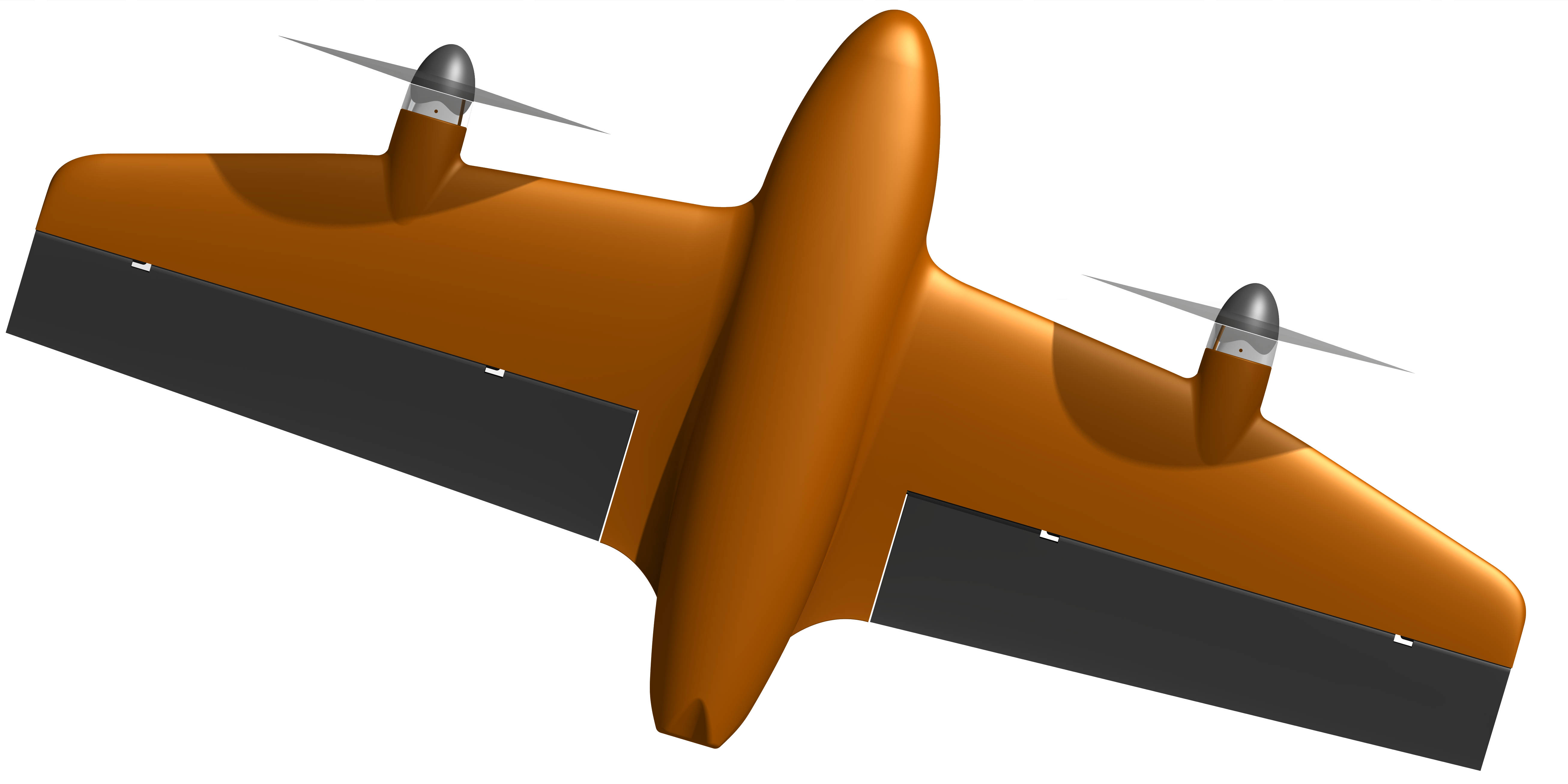}}};

\coordinate (O) at (-1.2,-1) ;
\coordinate (V) at (-2.9,-1) ;
\coordinate (A) at (0.25,0) ;
\coordinate (B) at (98:1.5) ;
\coordinate (V1) at (35:1) ;
\coordinate (C) at (52:3) ;
\coordinate (X1) at (52:1) ;
\coordinate (D) at (-0.2,3) ;

\draw[>=stealth,->] (O) -- (V) node[left] {$V$} ;

\end{tikzpicture}

%% file: aos.tex
\begin{tikzpicture}

\node[inner sep=0pt] (cyclone) at (-0.9,-1.1)
    {\includegraphics[width=.25\textwidth]{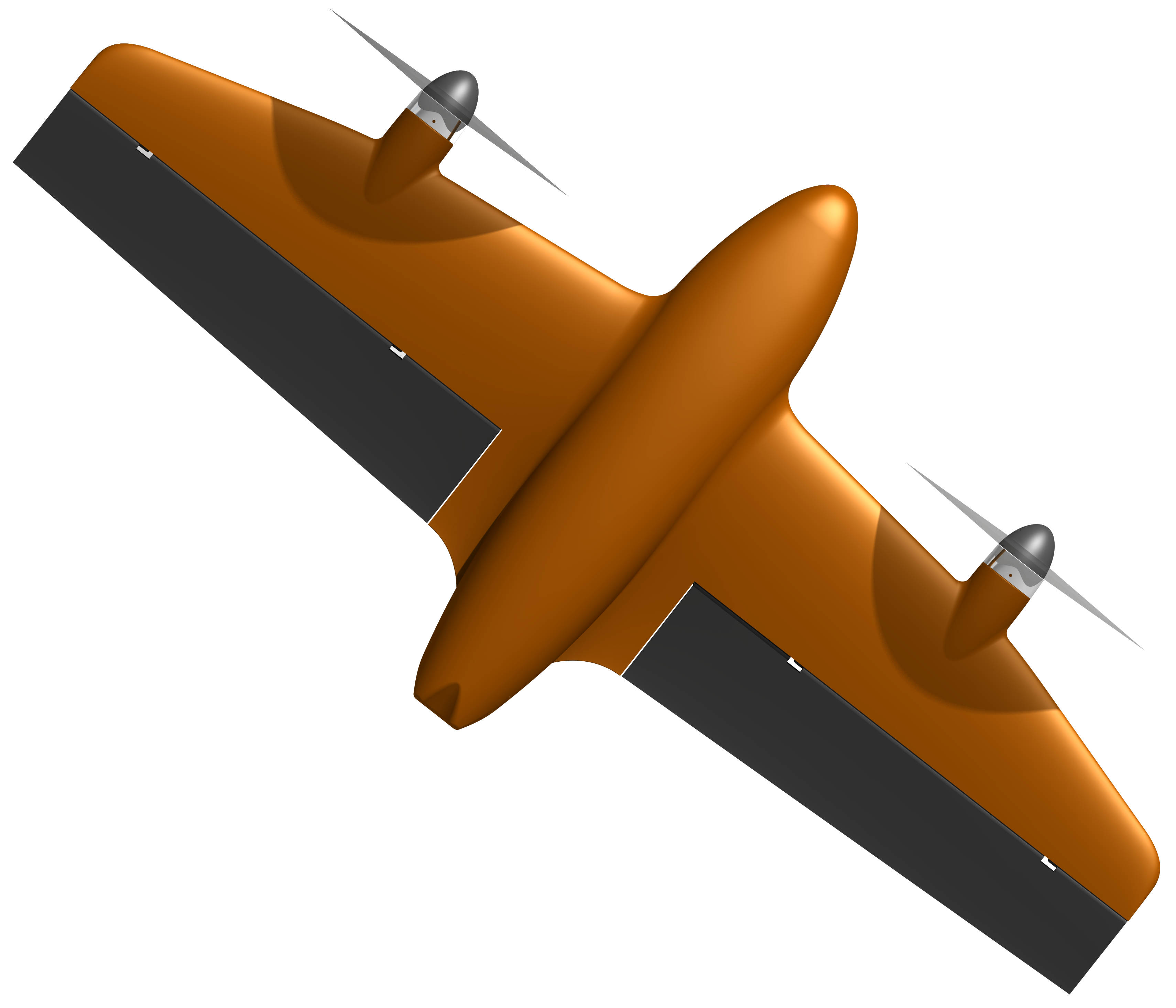}};

\coordinate (O) at (0.1,0.1) ;
\coordinate (O1) at (-0.2,0.1) ;
\coordinate (A) at (3,1) ;
\coordinate (B) at (35:3.5) ;
\coordinate (V1) at (35:1) ;
\coordinate (C) at (52:3) ;
\coordinate (X1) at (52:1) ;
\coordinate (D) at (-0.2,3) ;

\draw[->] (O) -- (C) node[right] {$-Z_B$} ;
\draw[>=stealth,->, thick] (0.1,-0.05) -- (B) node[draw=none,fill=none,right] {$V$};

\node[right] at (D)  {$X_N$};

\draw[->] (X1)  arc (53:35:1.2) ;
\node[] at (42:1.6) {$\beta$};
\draw[->] (O1) -- (D) ;
\draw[->] (0.22,-0.25) -- ++(-38:2) node[right] {$Y_B$} ;
\end{tikzpicture}